\documentclass{article}

\usepackage[preprint]{neurips_2021}

\usepackage[utf8]{inputenc} % allow utf-8 input
\usepackage[T1]{fontenc}    % use 8-bit T1 fonts
\usepackage{hyperref}       % hyperlinks
\usepackage{url}            % simple URL typesetting
\usepackage{booktabs}       % professional-quality tables
\usepackage{amsfonts}       % blackboard math symbols
\usepackage{amsmath}
\usepackage{nicefrac}       % compact symbols for 1/2, etc.
\usepackage{microtype}      % microtypography
\usepackage{xcolor}         % colors
\usepackage{graphicx}       % figures
\usepackage{booktabs}
\usepackage{placeins} 
\usepackage{amsmath}
\usepackage{subcaption}

\DeclareMathOperator*{\argmax}{arg\,max}

\title{Uncertainty evaluation of segmentation models \\ for Earth observation}

\author{%
  Mélanie Rey \\
  Google DeepMind \\
  \texttt{melanierey@google.com} \\
  \And
  Andriy Mnih \\
  Google DeepMind \\
  \texttt{amnih@google.com} \\
  \And
  Maxim Neumann \\
  Google DeepMind \\
  \texttt{maximneumann@google.com} \\
  \And
  Matt Overlan \\
  Google DeepMind \\
  \texttt{moverlan@google.com} \\
  \And
  Drew Purves \\
  Google DeepMind \\
  \texttt{dwpurves@google.com} \\
}

\begin{document}

\maketitle

\begin{abstract}

This paper investigates methods for estimating uncertainty in semantic segmentation predictions derived from satellite imagery. Estimating uncertainty for segmentation presents unique challenges compared to standard image classification, requiring scalable methods producing per-pixel estimates. While most research on this topic has focused on scene understanding or medical imaging, this work benchmarks existing methods specifically for remote sensing and Earth observation applications. Our evaluation focuses on the practical utility of uncertainty measures, testing their ability to identify prediction errors and noise-corrupted input image regions. Experiments are conducted on two remote sensing datasets, PASTIS and ForTy, selected for their differences in scale, geographic coverage, and label confidence. We perform an extensive evaluation featuring several models, such as Stochastic Segmentation Networks and ensembles, in combination with a number of neural architectures and uncertainty metrics. We make a number of practical recommendations based on our findings.
\end{abstract}

\section{Introduction}

Estimating prediction uncertainty for deep learning models is a complex yet crucial task for their application in complex, real-world domains. Uncertainty quantification is important for two primary reasons: first, some fields, such as medical science, require high-confidence predictions for practical use, and second, quantifying uncertainty offers a powerful method for improving models by highlighting their weaknesses and potential data issues. Uncertainty estimation for remote sensing and Earth observation has gained momentum recently \citep{Singh_2024, Lang_2022}; however, a systematic evaluation and comparison across different architectures and datasets is still lacking.

There are two main types of uncertainty to consider when making predictions using a model. The \emph{aleatoric} uncertainty corresponds to the noise inherent to the data-generating process when the output is not a deterministic function of the input; this is orthogonal to any modeling considerations. The \emph{epistemic} uncertainty is the uncertainty about the model class and its parameter setting used to produce the outputs from the inputs; it is sometimes referred to as the model uncertainty. In this work, we do not aim to quantify these uncertainty types separately, as doing so has been shown to be difficult in practice \citep{kahl2024values}. Instead, we focus on metrics that evaluate the overall uncertainty and answer pragmatic questions for practitioners:
(1) Can we detect when the model will make inaccurate predictions? (2) Can we identify corrupted data? (3) Can the model's performance be improved through (post-hoc) uncertainty estimation? While these questions do not encompass the full scope of what uncertainty should ideally capture (e.g., ambiguous examples or input distribution shifts), they provide a solid starting point for exploring model uncertainty in practical scenarios. Our analysis leads to the following main findings:

\begin{itemize}
\item Common uncertainty estimation methods demonstrate limited performance in the task of identifying misclassified pixels at test time. However, these methods are substantially more effective at identifying poorly segmented images when evaluated at the image level.
\item The choice of model architecture significantly impacts uncertainty estimation; we find that Vision Transformer-based models substantially outperform convolutional models at identifying segmentation errors.
\item Since performance varies across architectures and datasets, we recommend the practice of setting aside a dedicated ``uncertainty test set'' for robust evaluation.
\item We show that Stochastic Segmentation Networks (SSN), which use a Gaussian latent layer to model correlation between labels at different locations, work well with the Transformer architecture. They deliver improved segmentation performance in both the noise-free setting and when training with noisy inputs, compared to the standard Transformer-based models without latent variables.
\end{itemize}

\section{Related work}

Estimating the uncertainty of deep learning model predictions is an active area of research, particularly in probability calibration and uncertainty modeling.

The first research stream investigates whether the confidence scores produced by a model, typically the output probabilities from a softmax layer, are calibrated, meaning they accurately reflect the true likelihood of correctness. Early work focused on developing reliable metrics for measuring calibration and exploring the impact of model architecture and capacity on miscalibration \citep{Guo2017_2, Nixon2020_3, Minderer_4}. Subsequent studies have also examined the challenge of maintaining calibration under dataset distribution shifts \citep{Ovadia2019_5, Cygert2021_6}. A variety of post-hoc and in-training methods have been proposed to improve model calibration \citep{Kumar2020_7, Mukhoti2020_8, Kull2019_9}.

The second stream, often inspired by Bayesian principles, focuses on directly modeling a model's uncertainty. Seminal methods in this category include Monte Carlo (MC) Dropout \citep{gal2016dropout}, which casts Dropout training as approximates Bayesian inference, and Deep Ensembles \citep{lakshminarayanan2017simple}, which leverage the diversity in ensemble member predictions to estimate uncertainty.

Extending these concepts to semantic segmentation introduces additional complexities, as uncertainty is no longer a single value for an image but a dense, pixel-wise map where spatial correlations are important. \citet{Kendall_2017} were among the first to systematically tackle this, adapting uncertainty techniques for dense prediction tasks. Much of the foundational work in this area has been conducted in the domain of medical imaging, where quantifying uncertainty is critical for clinical decision-making. Researchers in this field have successfully applied a range of techniques, including Bayesian neural networks, ensembles, and MC Dropout, to tasks like tumor and organ segmentation \citep{Ng2023_11, Hoebel2020_12, kwon2020_13}. The specific challenge of calibrating segmentation models has also been addressed, with studies investigating the impact of training procedures like the Dice loss \citep{Mehrtash_2019} and introducing new methods like selective scaling tailored for segmentation outputs \citep{Wang_2023}.

Despite active research on uncertainty estimation methods \citep{holder2021efficient, Obster2024, maag2024}, comprehensive benchmarks evaluating uncertainty estimation for segmentation are relatively rare. Notable exceptions include \citet{Ng_2022}, who evaluated MC Dropout, Deep Ensembles, and other methods on cardiac MRI datasets, and \citet{Aerts2020AnEO}, who explored different uncertainty quantification methods at both the pixel and image level for lung nodule segmentation. More recently, \citet{Buchanan2022_10} provided an extensive benchmark across a variety of datasets, and \citet{kahl2024values} proposed a structured framework for evaluating the practical value of uncertainty in segmentation based on predominant uncertainty applications such as failure detection and out-of-distribution detection. Whereas \citet{kahl2024values} focuses on image-level uncertainty evaluation, we consider pixel-wise metrics as a local spatial analysis is desirable in many remote sensing applications. 

In the domain of remote sensing, uncertainty estimation is an emerging but not yet standard practice. While some recent work has focused on related problems like out-of-distribution detection \citep{ekim202}, bias in map-derived regression coefficients \citep{lu2025regression} or ensemble learning for regression tasks \citep{Lang_2022, Lang_2023}, research on segmentation-specific uncertainty remains less common. Existing studies have explored Bayesian methods \citep{haas2021_14, Dechesne2021_15}, ensembles \citep{andersson2021model} and stochastic noise models in the logit space \citep{Pascual_2018}, though without modeling spatial noise correlation, a feature explored by \citet{monteiro_2020} in the context of medical imaging. More recently, conformal prediction has been explored as an alternative for providing uncertainty guarantees in Earth observation applications \citep{valle2023quantifying, Singh_2024}.

\section{How to assess uncertainty estimates?}
\label{sec:assessing_uncertainty}

Evaluation of uncertainty estimates presents a significant challenge due to the general unavailability of the ground truth. As a result, prior work has often focused on visualizations and theoretical properties rather than direct quantitative assessment \citep{Singh_2024}. To move beyond purely qualitative analysis, we propose a pragmatic evaluation framework focusing on the most common applications for uncertainty. We assess the quality of uncertainty estimates based on their utility in three key tasks: 1) identifying incorrect model predictions, 2) improving precision-recall on the segmentation task, and 3) detecting anomalies in the input images.

\paragraph{Identifying Segmentation Errors.}
Our primary evaluation assesses the ability of various uncertainty estimates to identify the model's segmentation errors on a test set. We frame this as a binary, pixel-wise classification task of distinguishing between the correctly classified pixels and the misclassified ones. For each pixel $i$, the corresponding uncertainty value $u_i$ is calculated and rescaled to the unit interval $[0,1]$. A pixel is then classified as an error if its uncertainty exceeds a given threshold $T$ (i.e., $u_i > T$). By varying this threshold across its full range, we generate an \emph{Uncertainty-Error Precision-Recall} (UE-PR) curve. In this context, perfect precision means every pixel classified as an error was indeed an error under the model, while perfect recall means every model error was successfully identified as such. This UE-PR curve, therefore, evaluates the quality of the uncertainty metric itself, largely decoupling the analysis from the underlying model's overall segmentation performance.

\paragraph{Post-hoc Performance Improvement.}
While the previously described task assesses the ability of an uncertainty estimate to identify model errors, a key practical application is to improve model performance by not making predictions on inputs for which the model is uncertain. To evaluate this use case, we compute a Precision-Recall (PR) curve for the original segmentation task, where predictions for pixels with uncertainty exceeding a threshold are excluded. This strategy creates a trade-off: not making a prediction necessarily reduces recall, but it can increase precision if the rejected pixel would have been misclassified. Crucially, this PR curve corresponds to the original segmentation task (which is multi-class classification), in contrast to PR curves for the preceding task, which involve the binary correct/incorrect classification problem. For multi-class segmentation, we report macro-averaged PR obtained by averaging precision and recall for individual classes. In contrast to methods like \citep{lee2020efficient} that require uncertainty-aware training, our evaluation framework is entirely post-hoc and applicable to any pre-trained model, regardless of its architecture or the loss function used to train it.

\paragraph{Identifying Noise-Corrupted Regions.}
Prediction uncertainty should ideally be useful not only for identifying  the model's mistakes but also for detecting atypical inputs. To evaluate this ability, we produce out-of-distribution inputs by adding spatially structured noise to them. We then assess the model at their ability to identify the corrupted/noisy input pixels at test time. This is formulated as a binary classification task, performance on which is evaluated using \emph{Uncertainty-Noise Precision-Recall} curves obtained by varying the uncertainty threshold as was done above for identifying segmentation errors. When computing precision, we exclude uncorrupted, incorrectly classified (in the original segmentation task) pixels for which the model predicts high uncertainty, as we do not want to penalize models for having high uncertainty on incorrectly predicted pixels.

\section{Uncertainty estimation}

We focus on methods that are either designed for segmentation, or are computationally efficient and applicable pixel-wise.

\paragraph{Functions of Estimated Probabilities.}
A straightforward approach to uncertainty estimation is to derive metrics directly from the model's output class probabilities. For each pixel $i$, we evaluate two functions of its predicted probability vector, $\boldsymbol{\hat{p}_i} = (\hat{p}_{i1}, \dots, \hat{p}_{iK})$, where $K$ is the number of classes.

\begin{itemize}
    \item \textbf{Normalized Entropy:} The first metric is the Shannon entropy of the predicted class distribution, $-\sum_k \hat{p}_{ik} \log \hat{p}_{ik}$. We normalize this value by the maximum possible entropy for a $K$-class distribution, $\log K$, to yield a final uncertainty score in the $[0,1]$ range.
    \item \textbf{Maximum Probability:} The second metric, often called the \emph{maxprob score}, is a simple yet effective baseline from the literature \citep{blum2019fishyscapes}. It is the maximum value in the probability vector, $\max_k \{\hat{p}_{ik}\}$. Since high probability corresponds to low uncertainty, we define the metric as $1 - \max_k \hat{p}_{ik}$. This ensures that higher values correspond to higher uncertainty, making it directly comparable to the entropy score.
\end{itemize}

\paragraph{Ensemble Methods.}
We construct an ensemble by independently training several models using the same hyperparameters but different random seeds for initialization. After training, we combine the models predictions for each pixel using two standard approaches:
\begin{itemize}
    \item \textbf{Ensemble Product} We average the logit vectors from the models before applying the softmax function to obtain class probabilities. 
    \item \textbf{Ensemble Mixture} We average the class probability vectors from the models as done by \citet{lakshminarayanan2017simple}. 
\end{itemize}
\noindent For ensembles we consider the following uncertainty metrics:
\begin{itemize}
    \item \textbf{Inter-model Variance:} We compute the variance of the predicted class probabilities for each pixel across the models in the ensemble, quantifying the disagreement among the ensemble members. 
    \item \textbf{Functions of the Mean Prediction:} This approach first calculates the ensemble's single, combined prediction (either via the \emph{product} or \emph{mixture} method) and then applies the functions described above (i.e., Normalized Entropy and Maximum Probability) to this resulting probability vector \citep{Mehrtash_2019, Hoebel2020_12}.
\end{itemize}

\paragraph{Stochastic Segmentation Networks.} \label{ssn_intro}
Whereas ensemble methods emulate Bayesian averaging by introducing random variation between different models, stochasticity can be represented explicitly in the form of a random layer in the model: a latent variable, e.g.~Gaussian, is introduced for each output dimension to explicitly model aleatoric uncertainty.
While this approach seems natural for regression, the case of classification is less straightforward: we cannot add noise directly to the predicted probabilities without leaving the probability simplex. \cite{Kendall_2017} proposes instead to obtain stochastic probability predictions by adding noise in the logit space, that is, to the inputs of the softmax that produces the output probabilities.
\citet{monteiro_2020} extends this approach to take into account the spatial context of the segmentation task by modeling the correlations between the latent variables, resulting in Stochastic Segmentation Networks. We describe this approach below.

Consider an input image $x$ with corresponding label mask $y$, with their pixels indexed using $i=1,\dots,S$. The simplest approach would be to make the logit vectors $z_i=(z_{i1},\dots,z_{iK})$ noisy but with the noise across pixel locations being independent:
\begin{equation}
z_i | x \sim\mathcal{N}(\mu_i(x), \sigma^2_i(x)), \: p_i = \mathrm{softmax} (z_i), \forall i,
\end{equation}
where $\mu_i(x), \sigma_i(x) \in \mathbb{R}^K$ are the outputs of a neural network. This equation incorporates uncertainty modeling in the segmentation task such that conditioned on the input $x$ the noise is independent across pixels. To obtain spatially consistent predictions however we need a noise model over the whole segmentation mask where noise across pixels is correlated. Let $z$ denote the (flattened) random vector of logits for all pixels: $z=(z_{11},\dots, z_{1K}, \dots, z_{S1}, \dots, z_{SK})$. We can model the joint distribution as:
\begin{equation} \label{mvg}
z | x \sim\mathcal{N}(\mu(x), \Sigma(x)), \; \;
\mu(x) \in \mathbb{R}^{SK}, \;
 \Sigma(x) \in \mathbb{R}^{SK \times SK}, 
\end{equation}
and we still assume that the probabilities $p_i$ are independent across pixels given the logits $z$.
Having introduced a noise variable $z$ we obtain a latent variable model that is fully specified by the joint likelihood (conditioned on the image) and can be trained using maximum likelihood:
\begin{equation}\label{likelihood}
p(y, z |x) = p(y | z)p(z |x)= \textstyle{\prod_i} p(y_i |z) p(z|x),
\end{equation}
where the second equality comes from the modeling assumption that conditioned on the logits $z$ the output is independent of the input image\footnote{This can be contrasted to the likelihood in a classic deterministic segmentation model: $p(y|x)=\prod_j p(y_i|x)$.}.
The likelihood defined in Eq.~\eqref{likelihood} cannot be computed in closed form but can be estimated by using samples of the Gaussian logits. This procedure is efficient because generating multiple samples requires just a single forward pass through the network to compute the parameters of the Gaussian in Eq.~\ref{mvg}. 
As the number of parameters in the covariance matrix $\Sigma$ is quadratic in the number of random variable dimensions, learning it directly is infeasible for all but the smallest image sizes. \citet{monteiro_2020} therefore uses a low-rank decomposition of the matrix to dramatically reduce the number of parameters involved:
\begin{equation}
    \Sigma = P P^T  + D,
\end{equation}
where $D$ is a diagonal matrix and $P \in \mathbb{R}^{SK \times R}$ for a chosen hyperparameter $R$, representing the rank of the parametrization. We typically choose values of $R \sim 10$, much smaller than $SK$ which makes learning feasible in practice.

\begin{figure}
    \centering
    \includegraphics[width=0.88\linewidth]{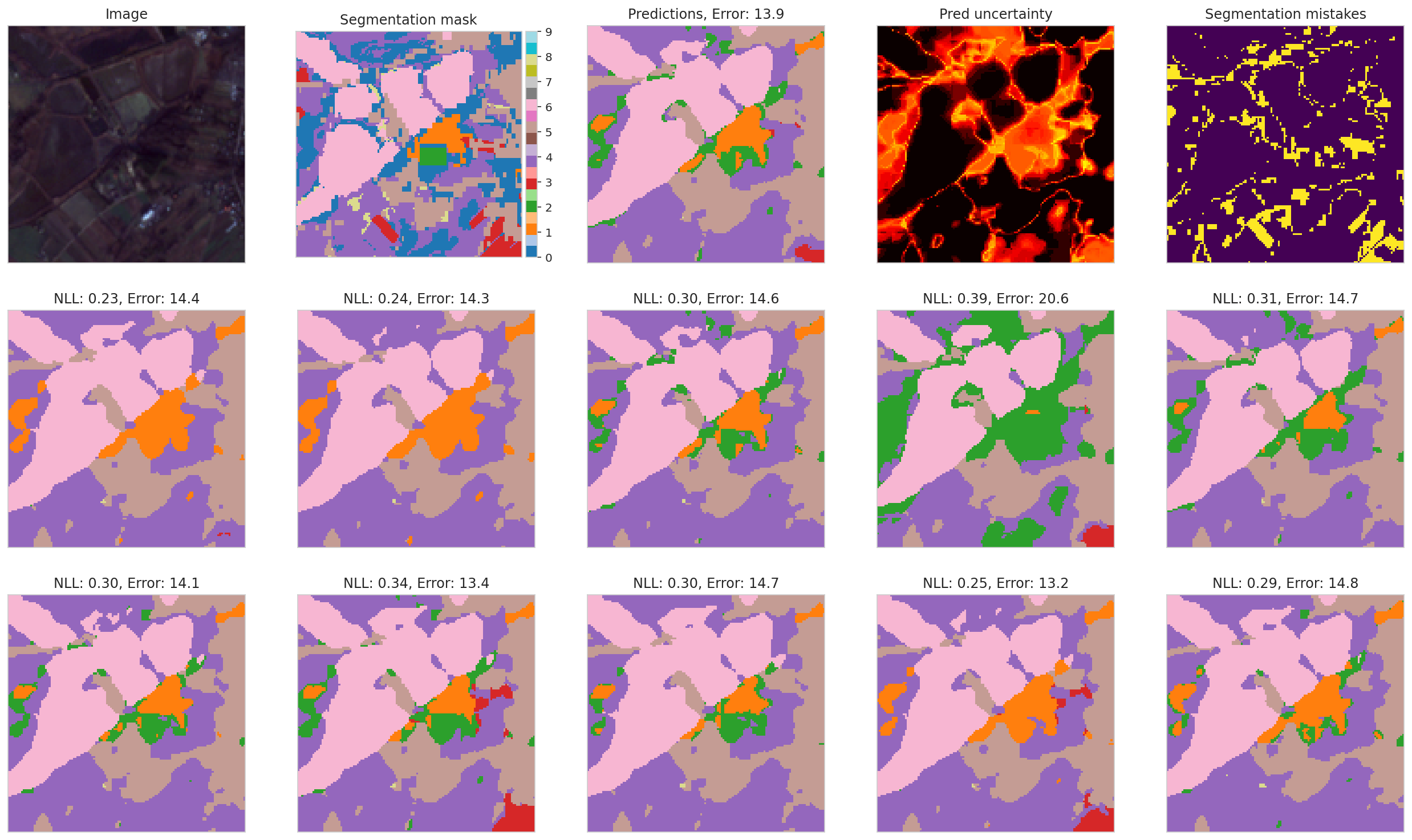}
    \caption{Results from SSN on ForTy. Top row from left to right: input image, segmentation labels, predicted segmentation, model uncertainty (Sampled Categorical Variation), segmentation mistakes. Middle and Bottom rows: alternative predictions sampled from the model. Note that the dark blue segments represent the background class and are ignored.}
    \label{ForTy_samples}
\end{figure}

We make three key changes compared to \citet{monteiro_2020}, which we found led to better model performance:
\begin{itemize}
    \item Instead of adding an additional layer to obtain the extra parameters necessary for the Gaussian logit distribution we simply made the output layer wider. More specifically, we extend the dimension of the embedding directly preceding the Gaussian layer from $(H, W, K$) to  $ (H, W, (2+R) K)$, where $H, W$ denote the height and width of the segmentation mask, respectively. 
    \item We applied a small fixed scaling factor to $P$ and $D$ to improve the learning dynamics. 
    \item Whereas \citet{monteiro_2020} makes predictions using the mean of the logit distribution, $\hat{y}_i = \argmax_k \mu_i(x)$, we first estimate the per-pixel marginal conditionals from samples: $\hat{p}(y_i | x) = \frac{1}{M} \sum_{j=1}^M p(y_i|z^j)$ where $z^j \sim p(z | x)$ (i.i.d.). Given the estimated conditionals, we make the prediction by computing the argmax for each one individually: $\hat{y}_i = \argmax_k \hat p(y_i=k|x)$. In all our experiments we used $M=16$ samples, doing this improved the mIoU of 0.14 on average for PASTIS.
\end{itemize}
One advantage of stochastic models such as SSN is that we can sample the latents multiple times to compute uncertainty values. We evaluate two methods for estimating uncertainty for SSN. The first one is proposed in \citet{monteiro_2020}: it is the \textbf{Marginal Entropy} of the class distribution estimated for each pixel obtained by marginalizing over the latent:
\begin{equation}
- \sum_k \hat{p}(y_i =k| x) \log \hat{p}(y_i =k| x), \; \textrm{where} \; \hat{p}(y_i | x) = \frac{1}{M} \sum_{j=1}^M p(y_i|z^j).
\end{equation}
The second method, which we introduce here as \textbf{Sampled Categorical Variation}, leverages the unique advantage of a model that generates correlated noise: the ability to draw spatially coherent samples of alternative predictions. These samples avoid the ``salt and pepper'' noise effect typical of models that assume independent pixel noise. To calculate this metric, we first draw $M$ plausible segmentation masks from the model, by sampling from the model's logit distribution $M$ times and applying the pixel-wise argmax for each sample. We then compute the pixel-wise variability across these masks using the coefficient of unalikeability \citep{kader2007variability}: $1-\frac{1}{M}\sum_k n_{ik}$ where $n_{ik}$ is the number of samples for pixel $i$ of class $k$. One example of this measure is provided in Figure~\ref{ForTy_samples}.

\section{Experimental protocol}

\subsection{Datasets}
We conduct our evaluation on two distinct remote sensing datasets for land cover analysis: PASTIS and ForTy. Both datasets provide, among others, optical multi-spectral satellite image time series (SITS) from Sentinel-2 and land cover segmentation masks as targets. 
They were chosen to contrast a high-quality, regional dataset with a large-scale, global benchmark characterized by more variable label quality. All images have a size of  128$\times$128 pixels which corresponds to 1280 $\times$ 1280 meters at 10 meter resolution.

The \textbf{PASTIS} (Panoptic Agricultural Satellite TIme-Series) dataset comprises 2,433 multi-spectral temporally dense (up to 61 observations per year) image sequences taken over approximately one year \citep{garnot2021panoptic}. The annotations cover different regions in France and provide high-confidence ground truth on 18 agricultural crop classes, including common types like corn, sunflower, and various cereals (e.g., soft winter wheat, winter barley).

The \textbf{ForTy} (FORest TYpes) dataset consists of about 200,000 globally distributed samples \citep{jiang2025not}. The primary purpose of the dataset is to differentiate between key forest types (natural forests, planted forests, and tree crops) alongside other coarse land cover types (other vegetation, water, ice/snow, bare ground, and built areas). To achieve global coverage, ForTy integrates labels from over 20 public data sources, which results in variable label quality compared to PASTIS. The full benchmark is multi-modal, including Sentinel-1 synthetic aperture radar, next to climate and elevation data. However, to ensure a controlled comparison in our study, we use a simplified version containing only the seasonal aggregates of the 10 Sentinel-2 spectral bands.

\subsection{Models and Hyperparameter selection}

Our study evaluates three different neural architectures: U-TAE \citep{garnot2021panoptic}, UNET3D \citep{Cicek20163DUnet}, and the transformer-based TSViT \citep{Tarasiou_2023_CVPR}. We consider three model types that can be based on any of these architectures: 1) The base/deterministic/standard model that simply uses the architecture to output the segmentation class probabilities for each pixel. 2) An ensemble model consisting of five independently trained base models of the given architecture. 3) The SSN/Gaussian model described in Section~\ref{ssn_intro} that has a latent Gaussian just before the final softmax and that uses the underlying architecture to compute the parameters of the Gaussian.

For each architecture, we first found a good set of hyperparameters by performing a grid search with its base (i.e.~deterministic) model version, sweeping over the batch size, learning rate,  number of epochs, as well as architecture-specific parameters, averaging over three random seeds (see Appendix \ref{hyper_appendix} for the resulting choices).
We then used the found hyperparameter configuration for all other models (i.e.~SSNs and ensembles) using the same architecture. This was done to avoid providing an advantage to the more complex models, and 
in practice provides a slight advantage to the base models over SSNs. The hyperparameters unique to SSNs were selected based on a preliminary hyperparameter sweep and were kept fixed afterwards throughout all experiments.

Since PASTIS is a fairly small dataset, we performed the hyperparameter sweep by training on folds 1-3 and using the mean Intersection over Union (mIoU) on the fourth fold to identify the best configuration. Models were then retrained on folds 1-4 and evaluated on the 5th fold. This process yielded an optimal learning rate of $10^{-3}$ for all models when combined with the Adam optimizer and the cosine decay schedule. We found that a batch size of 128 performed well for all models. The optimal training duration, however, varied by architecture: longer training was required for models which take as input not only the satellite image time series but also the corresponding dates (850 epochs for TSViT, 600 for U-TAE), whereas the purely convolutional UNET3D model was already successfully trained after 200 epochs. For the larger ForTy dataset, we adopted the hyperparameters from \cite{jiang2025not} with the exception of batch size which was kept at 128 (instead of 64). All models on ForTy were trained for 40 epochs and we used a simple temporal masking augmentation method which randomly selects dates for zero masking.

\subsection{Noise corruption}
\label{sec:noise_corruption}

To evaluate model uncertainty in out-of-distribution (OOD) scenarios, we investigate two types of structured noise. The first, \emph{Object-Level Corruption}, utilizes instance-level annotations to corrupt entire objects within an image. The second, \emph{Elliptical Noise Artifacts}, introduces elliptically-shaped noise regions, making it applicable to datasets lacking object-level annotations. For both methods, corruption is implemented by adding Gaussian noise, with its intensity controlled by the noise variance. To simulate realistic input corruptions and prevent models from trivially identifying noise through scale variations, we add the noise before any data normalization is applied. The noise strength is specified as a multiplier on  the training set's per-channel standard deviations $\sigma_c$; for example, a noise level of 0.5 corresponds to adding zero-mean Gaussian noise with a standard deviation of $ 0.5 \times \sigma_c$ to channel $c$. For any corrupted pixel, the same noise level is applied consistently across all spectral channels and temporal observations, which ensures uniform information loss across all such dimensions. 
%\andriy{Does this make sense?} \melanie{Yes, it's good}

\begin{figure}[t] 
  \centering
  \includegraphics[width=0.4\textwidth]{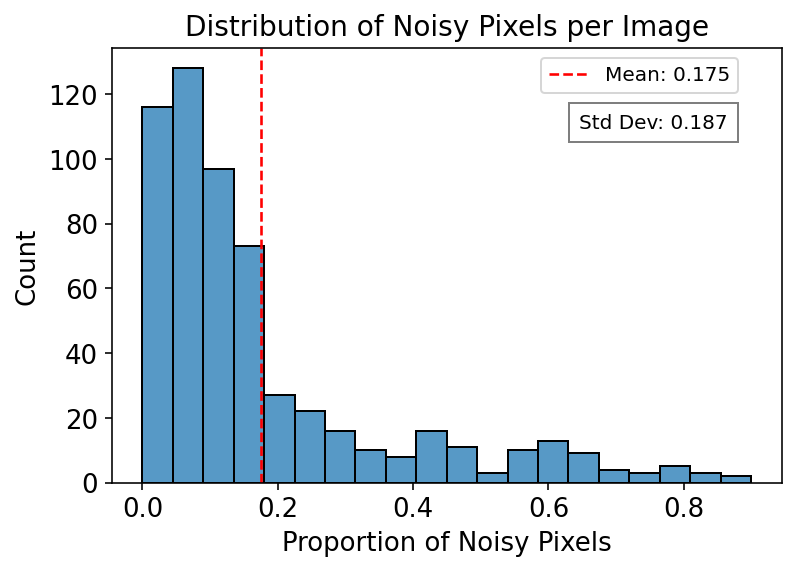}
  \includegraphics[width=0.4\textwidth]{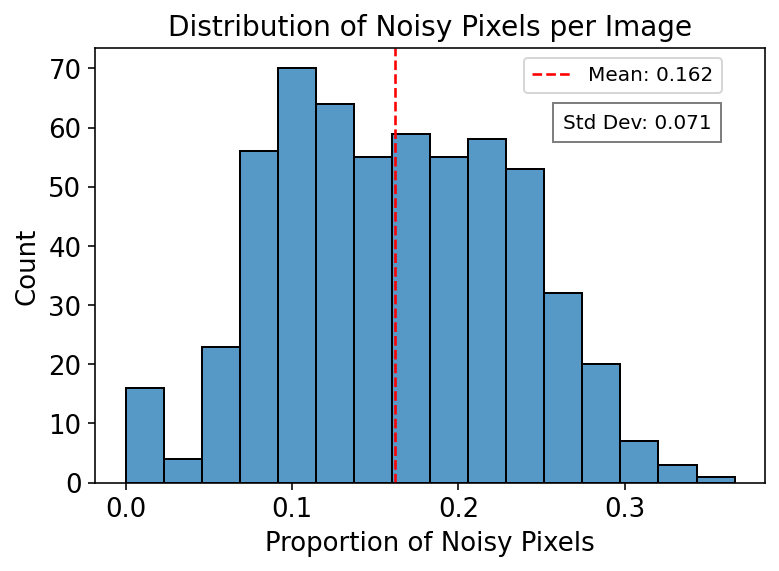}
  \caption{Distribution of the proportion of corrupted pixels per image, computed on the PASTIS test set. Left: For object-level corruption; Right: For elliptical noise.}
  \label{fig:noise_distrib}
\end{figure}

\paragraph{Object-Level Corruption.}
This approach utilizes the instance segmentation maps from the PASTIS dataset to create  OOD examples with realistic shapes. In each image, we select a random subset of object instances for corruption, with each instance having a selection probability of 0.2. To ensure meaningful perturbations, objects smaller than a 50-pixel threshold or those belonging to the ``void'' class are excluded from this process. This methodology results in an average corruption of 17.5\% of pixels per image, though there is substantial variability across images, as shown in the left panel of Figure~\ref{fig:noise_distrib}.

\paragraph{Elliptical Noise Artifacts.}
To be able to perform experiments on ForTy, which does not provide instance maps, we create artificial noise using elliptical object shapes. For each image, we randomly generate one or two ellipses, with the centers located uniformly at random, and their radii  drawn uniformly from the range $[20, 45]$ pixels. The right panel of Figure~\ref{fig:noise_distrib} shows the resulting distribution of the fraction of noisy pixels per image on PASTIS. It looks much less skewed than the distribution with the object-based noise because there is much less variability in the size/area of our elliptical ``objects'' compared to that of the real objects. We do not show the corresponding distribution for ForTy as it looks very similar.

\FloatBarrier

\section{Results on uncorrupted data}

\subsection{PASTIS}
Our segmentation results on the PASTIS dataset are provided in Table~\ref{tab:pastis_noiseless}. They are generally consistent with results in the literature \citep{garnot2021panoptic, Tarasiou_2023_CVPR, jiang2025not}, although our UNET3D model achieves a slightly higher mIoU. We observe two primary trends: ensemble models consistently yield the highest mIoU and Precision, and Stochastic Segmentation Networks outperform their deterministic counterparts for both the TSViT and U-TAE architectures. 

\begin{figure}[t]
  \centering
  \includegraphics[width=\columnwidth]{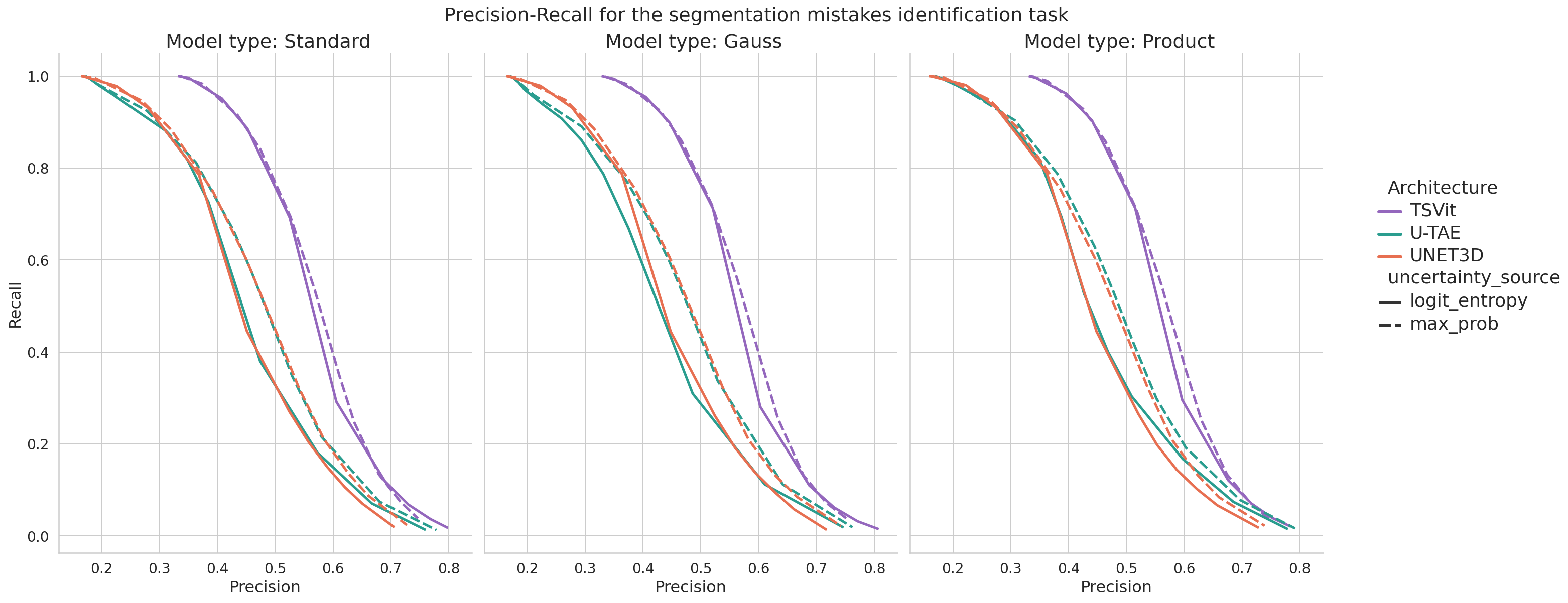}
  \caption{Uncertainty-Error Precision-Recall curves for identifying segmentation mistakes on PASTIS for different architecture / model type combinations. Left: Deterministic models. Middle: SSNs. Right: Ensembles.
  Transformers clearly outperform other architectures.}
  \label{all_wrong_2}
\end{figure}

We will now consider the task \emph{Identifying Segmentation Errors} from Section~\ref{sec:assessing_uncertainty}
 which requires estimating uncertainty about model predictions: distinguishing the pixels misclassified by the model from those classified correctly. Figure~\ref{all_wrong_2} shows the Uncertainty-Error Precision-Recall curves for different architecture/model/uncertainty measure combinations. We can see that of the two uncertainty measures, maxprob consistently outperforms entropy, though for TSViTs the performance gap is small. \citet{maag2024} have also observed that maxprob performed well, and in particular outperformed entropy, at pixel-wise uncertainty evaluation tasks even though it was insufficient for OOD detection. Curiously, there is very little difference in performance between different model types. In particular, despite their superior segmentation performance in Table~\ref{tab:pastis_noiseless}, ensemble models show no significant improvement over standard models on this task. This suggests that while the diversity of model predictions in  ensembles is sufficient to boost mIoU, it is less effective at improving uncertainty estimates for identifying prediction errors. 

The most prominent pattern in these results is the clear superiority of the Transformer architecture over the other two, for all model/uncertainty metric combinations. This finding is consistent with the observation by \citet{Minderer_4} that Vision Transformers are better calibrated than older architectures based on ResNets, such as SimCLR. More generally, on this task, the choice of architecture turns out to have far larger effects on performance than the choice of the model or uncertainty measure. 

\begin{figure}[t]
  \centering
  \includegraphics[height=160pt]{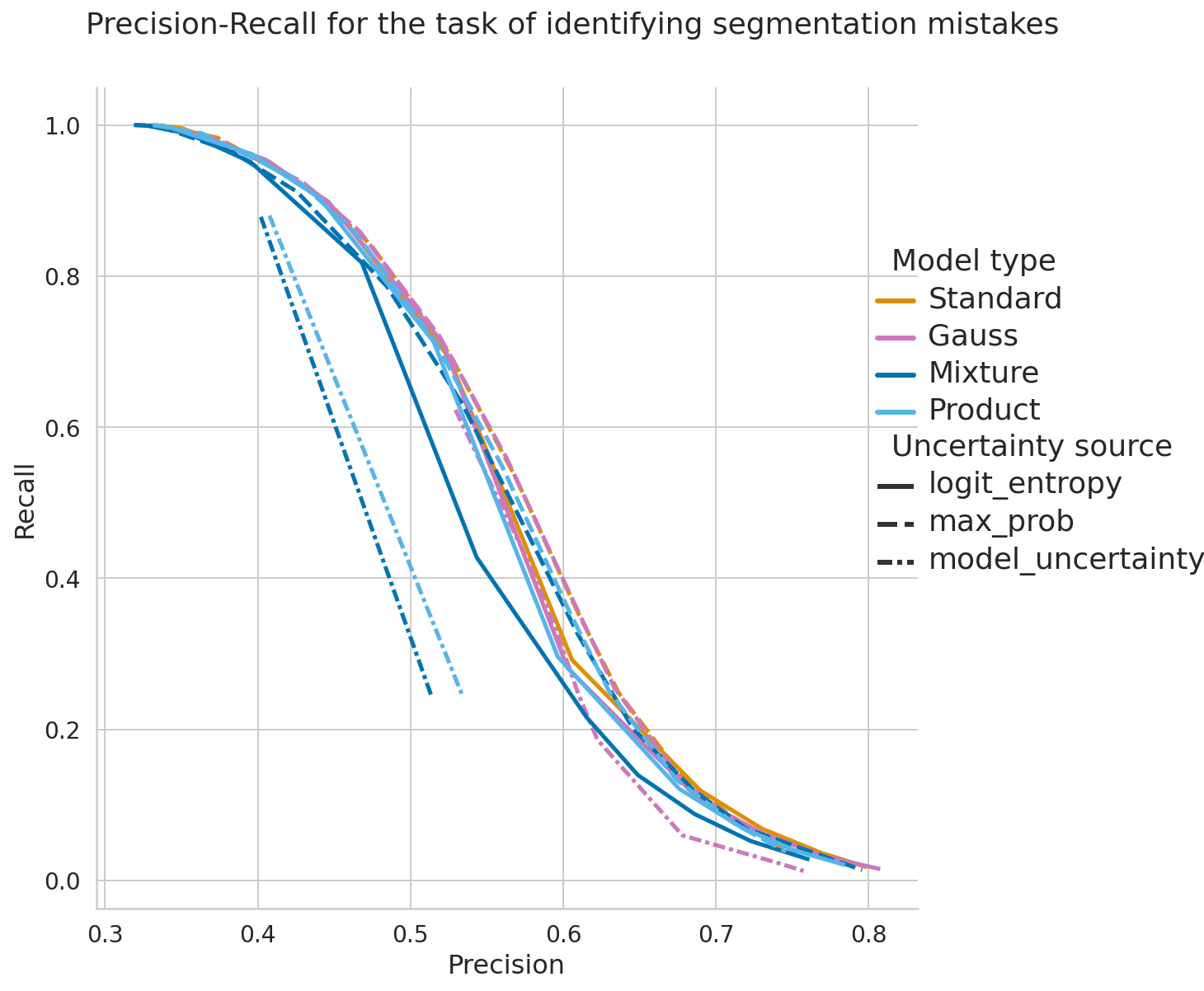}
  \includegraphics[height=160pt]{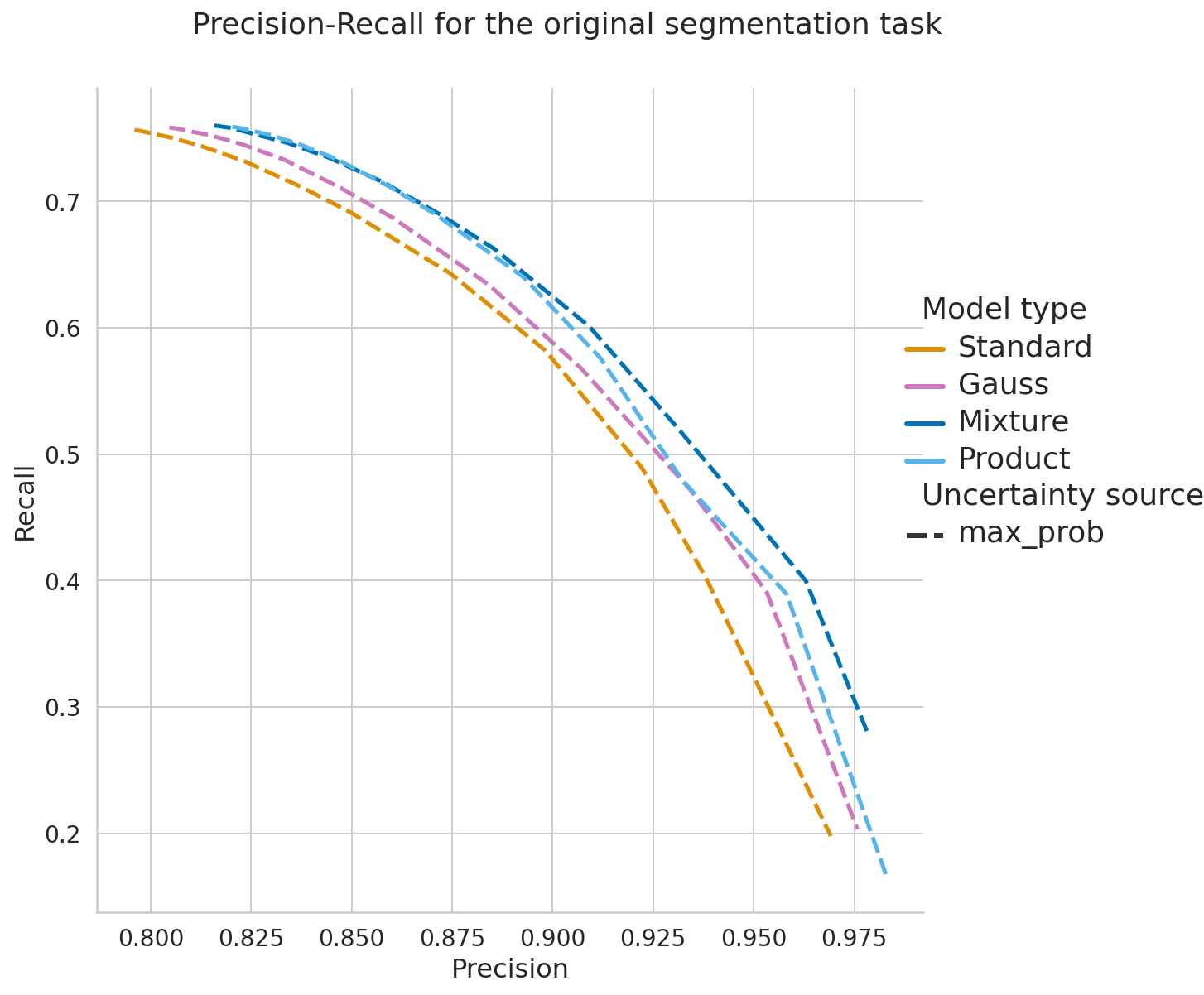}
  \caption{TSViT on PASTIS. Left: Uncertainty-Error Precision-Recall curves for the task of identifying segmentation mistakes. Right: Precision-Recall curves for the original segmentation task when using uncertainty estimates to reject pixels.}
  \label{pastis_dome}
\end{figure}

Given the superior performance of the TSViT architecture, we summarize the results for it in the left panel of Figure~\ref{pastis_dome}, also including new results for the ensemble mixtures as well as a model-specific uncertainty measure, which for ensembles is the inter-ensemble variance and for SSNs is the sampled categorical variation described in Section \ref{ssn_intro}. We see that the best performance is achieved by the SSN and the standard models using the maxprob uncertainty measure, though the two ensemble models using the same measure perform only slightly less well. The model-specific measures performed quite poorly on this task, especially for ensembles.

We now consider the Post-hoc Performance Improvement task from Section~\ref{sec:assessing_uncertainty}, which uses uncertainty estimates to boost segmentation performance. This task takes into account both segmentation performance and uncertainty estimation quality, providing a combined evaluation.
 The right panel of Figure~\ref{pastis_dome} shows how different models using the TSViT architecture perform on it using maxprob as the uncertainty metric. 
We can see that ensembles, especially the ones using the mixture formulation, perform best at this task. SSNs are the runner-up, providing a good compromise between model size and performance, having only slightly more parameters than the standard models (1.01 times more in this particular case), while the ensembles have 5 times as many parameters.

\begin{figure}[t]
  \centering
  \includegraphics[width=\columnwidth]{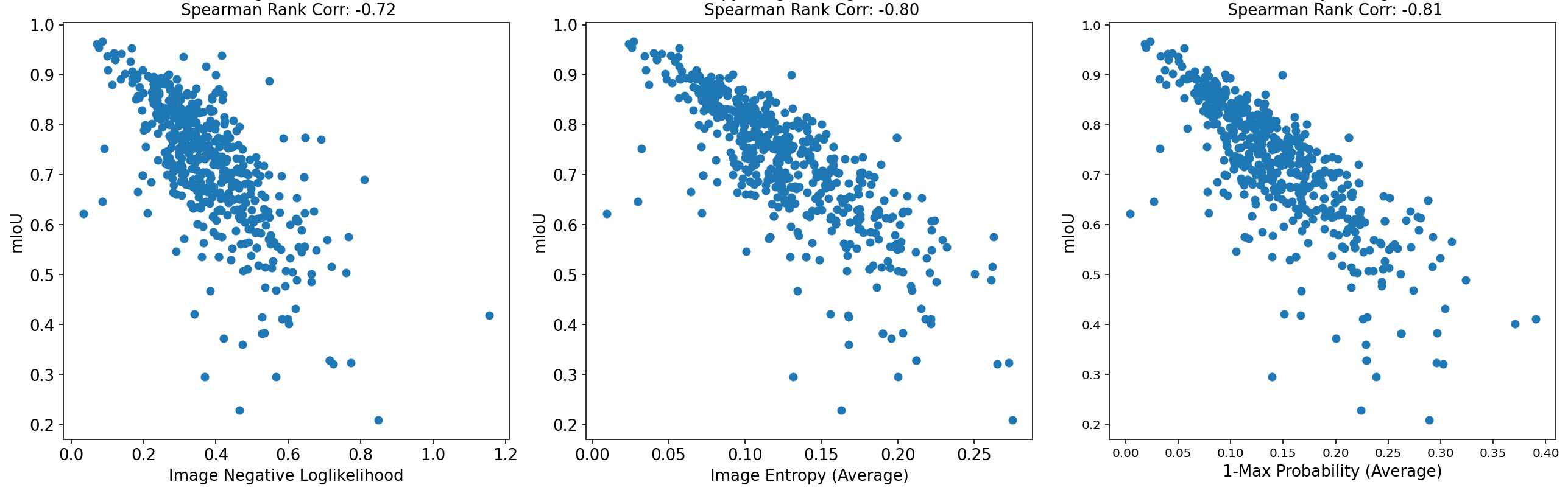}
  \includegraphics[width=\columnwidth]{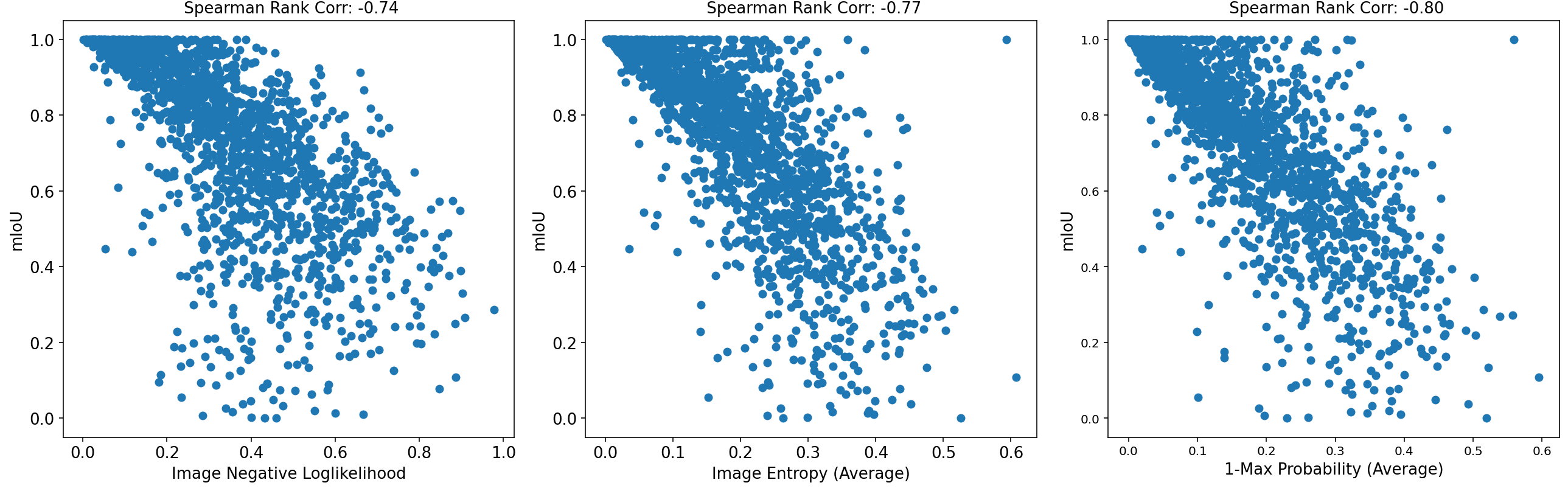}
  \caption{Image mIoU vs.~uncertainty measures for the TSViT SSN on: (top row) the PASTIS test set and (bottom row) 1600 randomly drawn images from the ForTy test set. Three measures of  uncertainty were considered: Left: Negative log-likelihood for the entire image estimated using 32 latent samples. Middle: (Marginal) pixel entropy averaged over the image. Right: 1-max probability averaged over the image.}
  \label{fig:miou_uncertainty}
\end{figure}

As prior work has explored image and segment-level uncertainty by aggregating pixel-wise values \citep{Mehrtash_2019, kahl2024values}, we now consider image-level uncertainty. This approach avoids dependence on segment annotations and object size bias. Instead of using a fixed threshold to classify images as well- or poorly-segmented, we examine the correlation between aggregated image-level uncertainty metrics and the mIoU score for the image. Figure~\ref{fig:miou_uncertainty} reveals a strong relationship, suggesting that uncertainty methods might be better at assessing the overall difficulty of an example than at identifying individual pixel errors. On this task, UNET3D's performance was comparable to that of TSViT. We also note that mean aggregation slightly outperformed median aggregation on the ForTy dataset, with the full results shown in Figures~\ref{forty_scatter_median_appendix} and  \ref{pastis_scatter_unet3d_appendix} in the Appendix.

\begin{table}[t!]
  \centering
  \caption{Model performance on four key metrics on the PASTIS test set. All scores are percentages (\%),  averaged over 3 random seeds. ``Gauss'' refers to the Stochastic Segmentation Network version of the model. ``Product'' refers to an ensemble that makes predictions by averaging the logits.}
  \label{tab:pastis_noiseless}
  \begin{tabular}{l cccc}
    \toprule
    \textbf{Model} & \textbf{mIoU} & \textbf{F1} & \textbf{Prec} & \textbf{Recall} \\
    \midrule
TSViT          & 64.77 & 77.32 & 79.60 & 75.65 \\
TSViT Gauss    & 65.35 & 77.76 & 80.46 & 75.84 \\
TSViT Product  & 66.23 & 78.46 & 82.03 & 75.90 \\
U-TAE          & 62.14 & 75.09 & 77.09 & 74.02 \\
U-TAE Gauss    & 62.50 & 75.45 & 74.60 & 77.26 \\
U-TAE Product  & 62.57 & 75.36 & 80.06 & 72.12 \\
UNET3D         & 64.31 & 76.86 & 79.10 & 75.49 \\
UNET3D Gauss   & 64.03 & 76.58 & 78.86 & 75.24 \\
UNET3D Product & 65.57 & 77.89 & 80.50 & 76.21 \\
    \bottomrule
  \end{tabular}
\end{table}

\begin{table}[h]
\centering
\caption{Model performance for on the ForTy test set. "Mixture" refers to an ensemble that makes predictions by averaging the probabilities.
}
\label{tab:forty_noiseless}
\begin{tabular}{lcccccc}
\toprule
\textbf{Model} & \textbf{mIoU} & \textbf{F1} & \textbf{Precision} & \textbf{Recall} \\
\midrule
TSVit          & 66.01            & 78.73       & 80.33              & 77.45           \\
TSVit Gauss    & 67.12            & 79.60       & 81.23              & 78.28           \\
TSVit Product  & 68.35            & 80.53       & 82.68              & 78.90           \\
TSVit Mixture  & 68.51            & 80.66       & 82.66              & 79.10           \\
\bottomrule
\end{tabular}
\end{table}

\subsection{ForTy}
We focused exclusively on TSViT on the ForTy dataset, as this architecture has substantially outperformed UNET3D and U-TAE in this setting in \citep{jiang2025not}. The segmentation performance of our models  reported in Table~\ref{tab:forty_noiseless} is slightly lower than that of \citet{jiang2025not}, due to our simplified experimental setup which uses only Sentinel-2 inputs (thus excluding Sentinel-1, elevation and climate data). Overall, our results on ForTy mirror those on PASTIS: the ensemble model performs best, followed by the SSN model.

\begin{figure}[t]
  \centering
  \includegraphics[height=160pt]{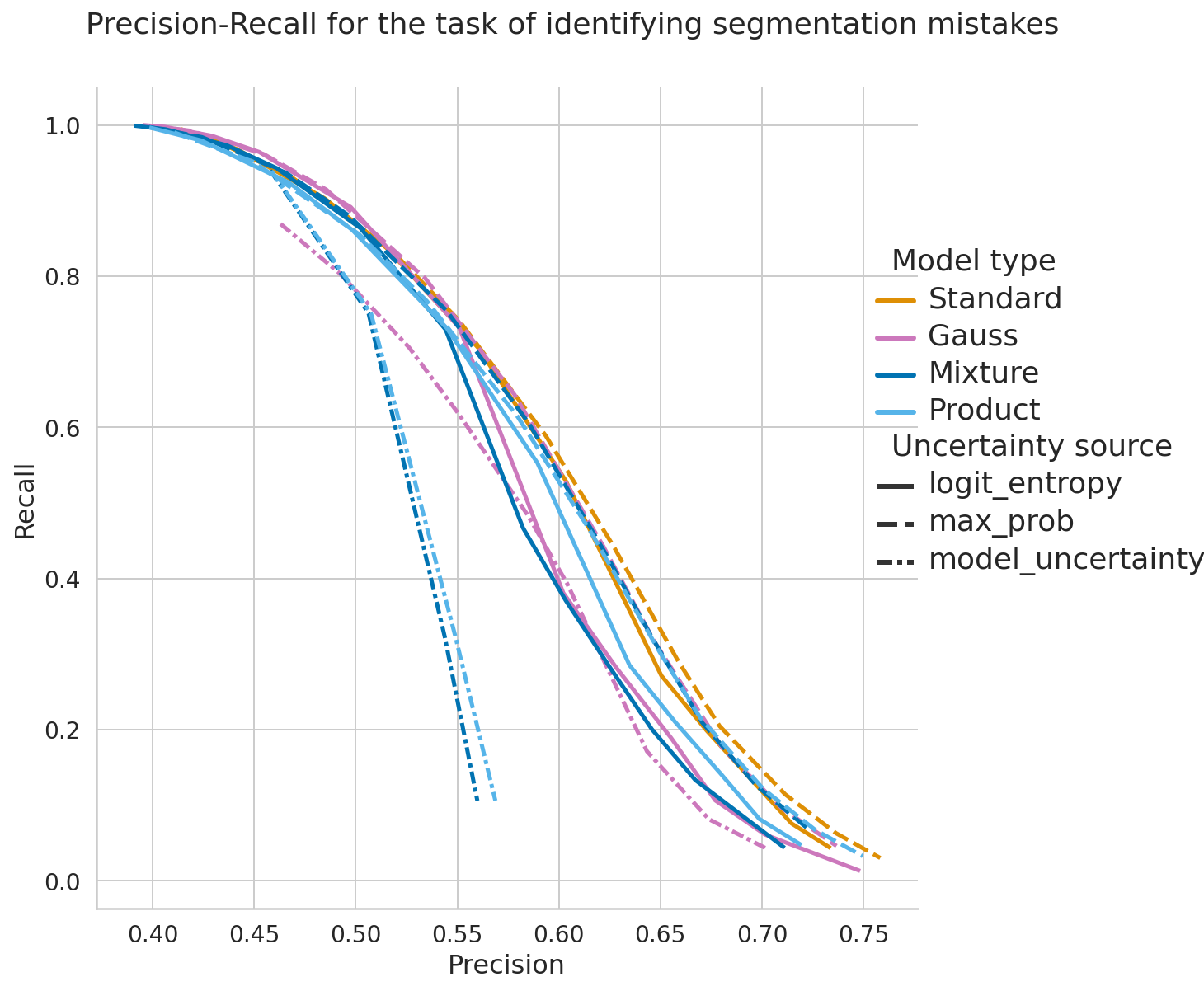}
  \includegraphics[height=160pt]{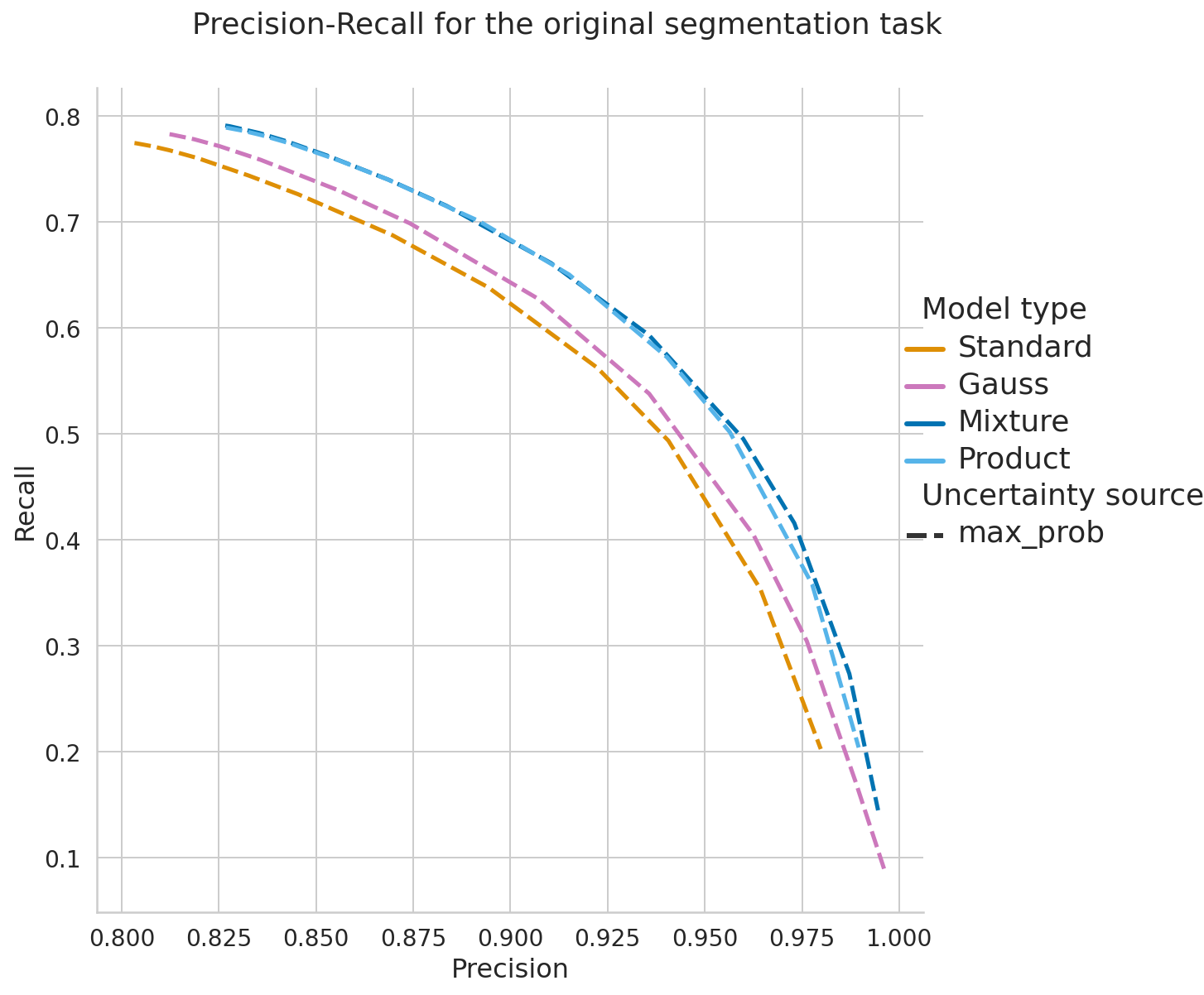}
  \caption{TSViT on ForTy. Left: Uncertainty-Error Precision-Recall curves for the downstream task of identifying segmentation mistakes. Right: Precision-Recall curves for the original segmentation task when using uncertainty estimates to reject pixels.}
  \label{fig:forty_pr_curves}
\end{figure}

The results on the two uncertainty evaluation tasks are shown in Figure~\ref{fig:forty_pr_curves}. The conclusions are very similar to those on PASTIS in both cases. On the task of identifying segmentation errors, presented in the left panel, maxprob consistently outperforms entropy, though the performance gap is relatively small in most cases. The model specific uncertainty measures once again underperform the other two. The standard model offers the best performance, with the SSN and the two ensembles very close behind.

For the uncertainty-boosted segmentation task, presented in the right panel of Figure~\ref{fig:forty_pr_curves}, the two ensembles perform best, with the mixture ensemble having the edge. As on PASTIS, the SSN model is the runner-up, providing a good compromise between performance and model size.

\FloatBarrier

\section{Results on noise corrupted data}

We now turn our attention to experiments involving inputs corrupted with structured noise.

\subsection{PASTIS}

\begin{figure}[h]
  \centering
  \includegraphics[width=0.72\textwidth]{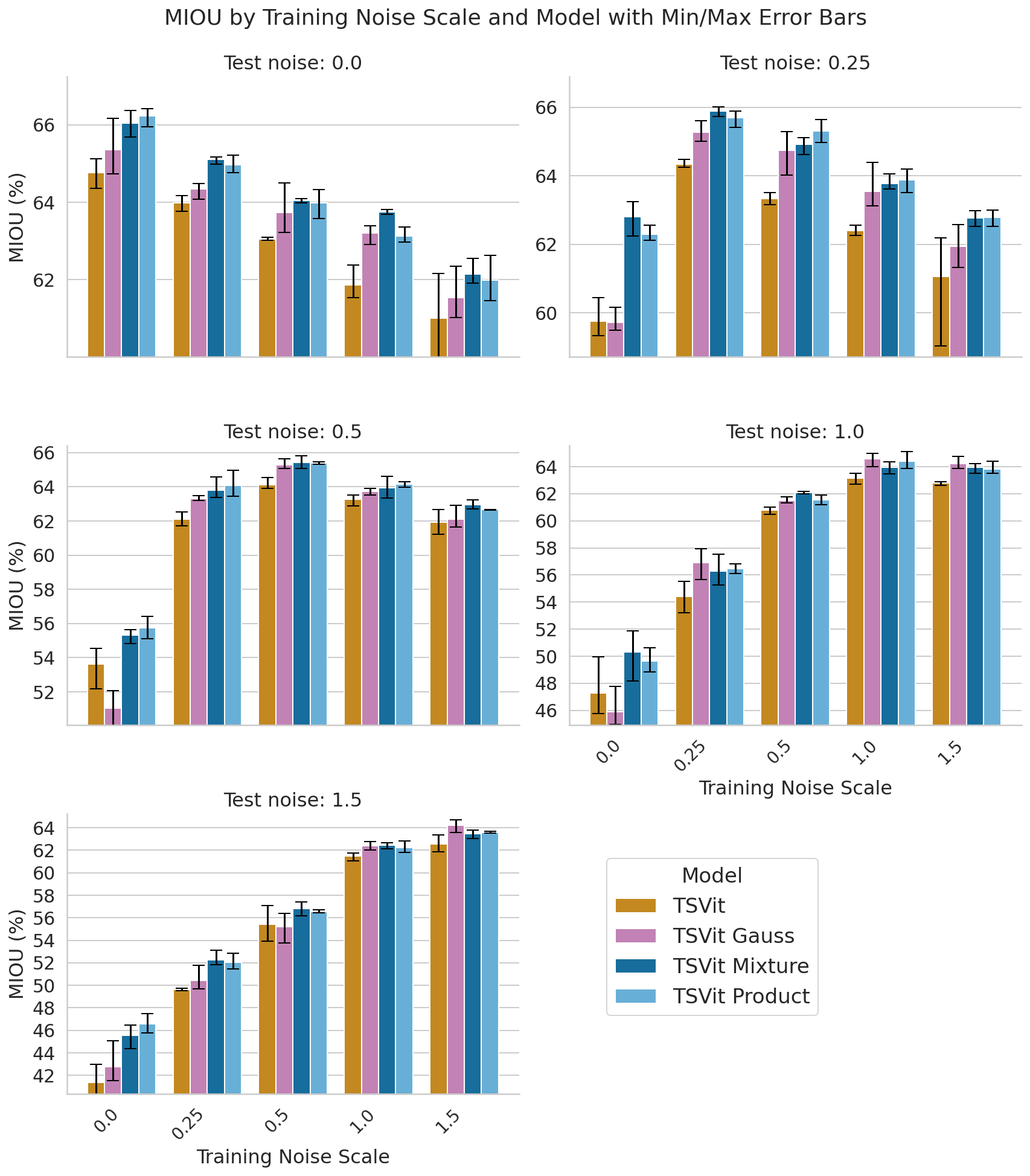}
  \caption{Object level noise on PASTIS: Segmentation performance for different train/test noise level combinations. We opted for varying y-axis scale to better emphasize the distinctions between models at identical evaluation noise levels.
}
  \label{fig:pastis_noise_objects}
\end{figure}

\begin{figure}[h]
  \centering
  \includegraphics[width=0.72\textwidth]{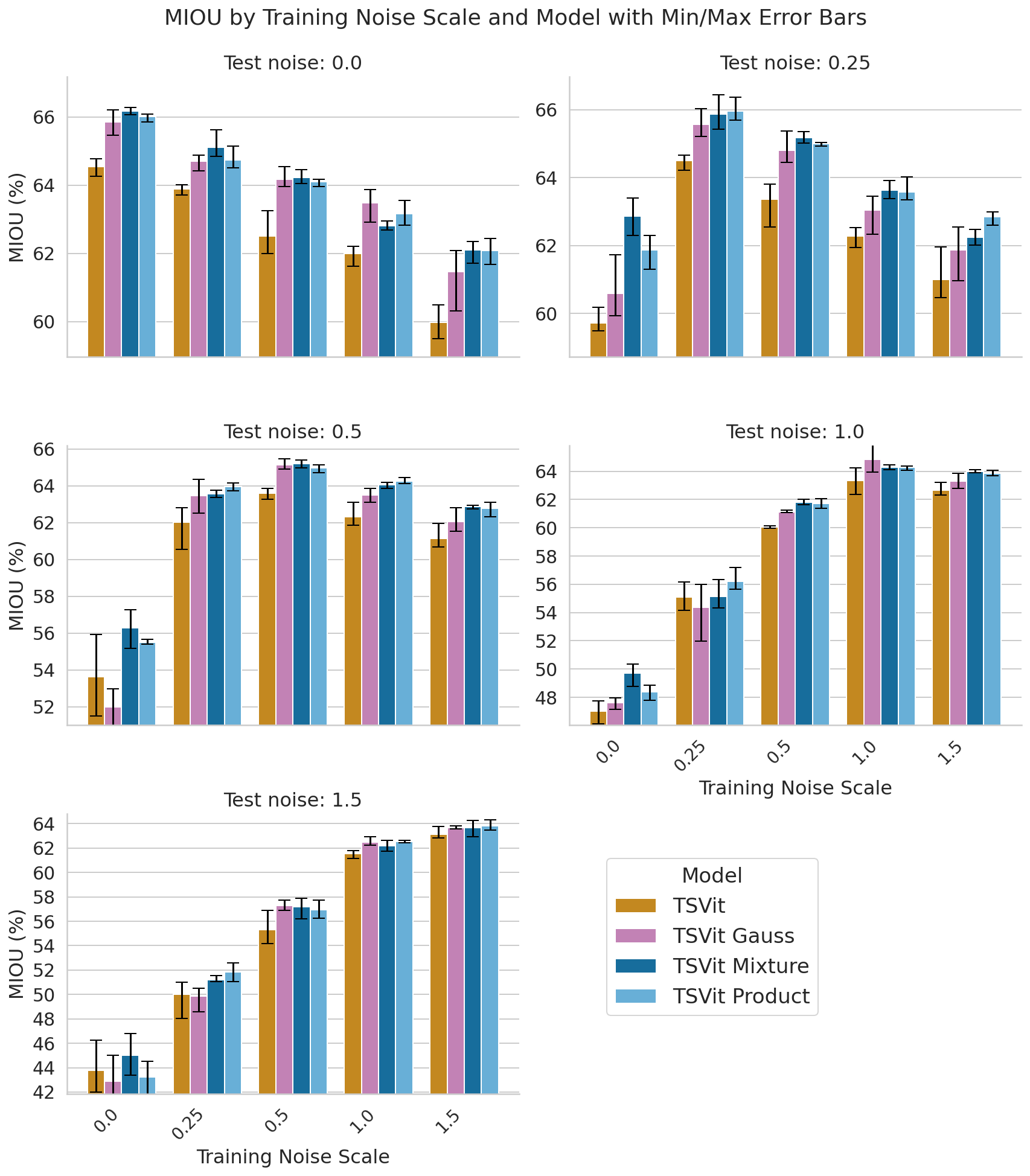}
  \caption{Elliptical noise on PASTIS: Segmentation performance for different train/test noise level combinations. We opted for varying y-axis scale to better emphasize the distinctions between models at identical evaluation noise levels.}
  \label{fig:pastis_noise_shapes}
\end{figure}

\paragraph{Segmentation performance}
Figures~\ref{fig:pastis_noise_objects} and \ref{fig:pastis_noise_shapes} show the segmentation performance results on PASTIS with object-instance noise and elliptical noise, respectively. 
As expected, we see that when training on clean data while evaluating on noisy data segmentation performance clearly degrades as the test noise level increases. For example, in Figure~\ref{fig:pastis_noise_objects} the standard TSViT model saw a steep decline in segmentation accuracy: its mIoU dropped from 64.8 with no test noise to 59.8 with the test noise level of 0.25. However, when trained and evaluated with the same noise level of 0.25, the model achieved an mIoU of 64.3. This rebound effect was even more pronounced with the SSN, for which the mIoU increased from 59.7 to 65.3. More generally, we observed that SSN models consistently benefited more from training with noise than standard models, sometimes even outperforming ensembles (e.g.~with a test noise level of 1.5).
Ensembles however, were the best performing models overall: not only were they more robust than other models to test time noise after training on clean data, but also performed comparably or better than the others when training in the presence of noise. For higher levels of noise, ensembles were sometimes slightly outperformed by SSNs, which were the second-best performing model class. Generally, all models performed best when the training and test levels of noise were the same. Interestingly, when evaluating with a non-zero level of test noise, training with the wrong level of noise resulted in better performance than training with no noise at all.

\begin{figure}[h]
\begin{subfigure}{0.485\textwidth}
  \centering
  \includegraphics[width=1.0\textwidth]{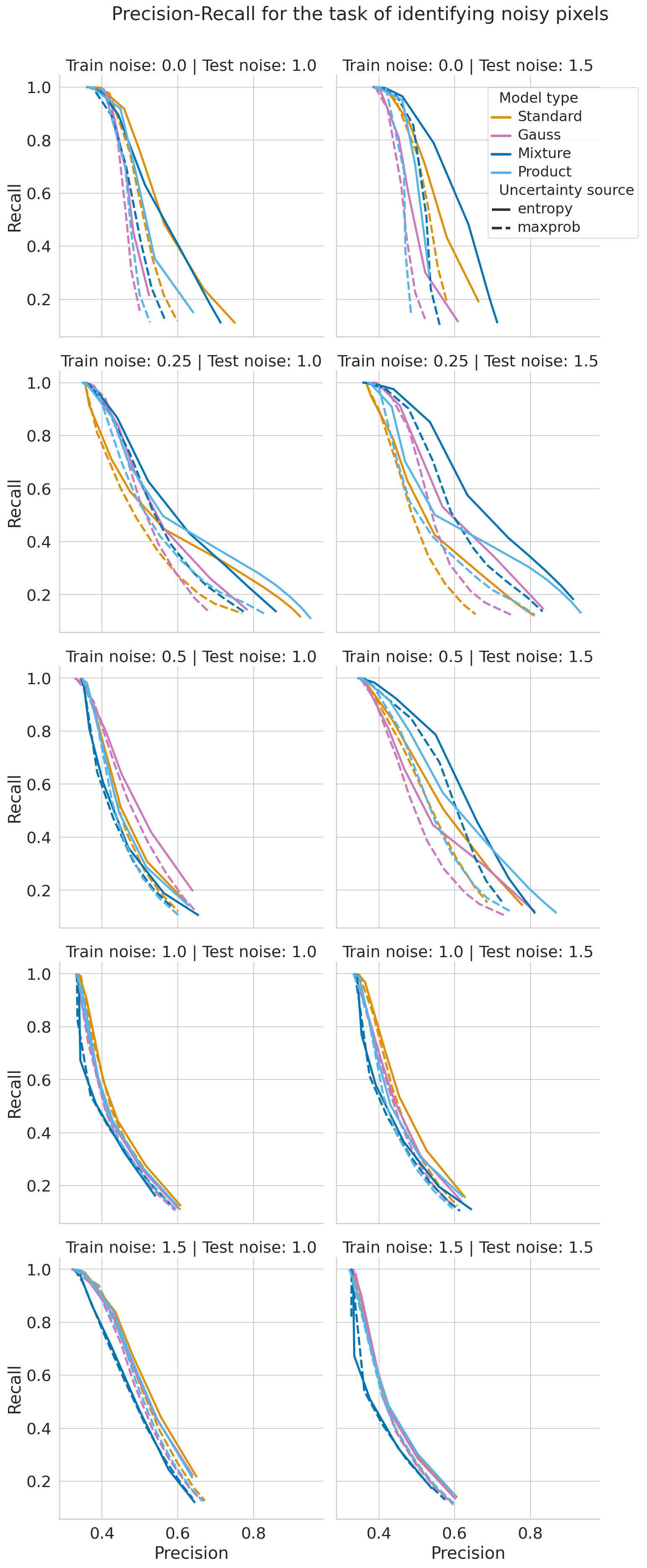}
  \caption{Object level noise on PASTIS. See Figure \ref{pastis_object_noise_appendix} in Appendix for further results.}
  \label{pastis_object_noise}
\end{subfigure}
\hfill
\begin{subfigure}{0.485\textwidth}
  \centering
  \includegraphics[width=1.0\textwidth]{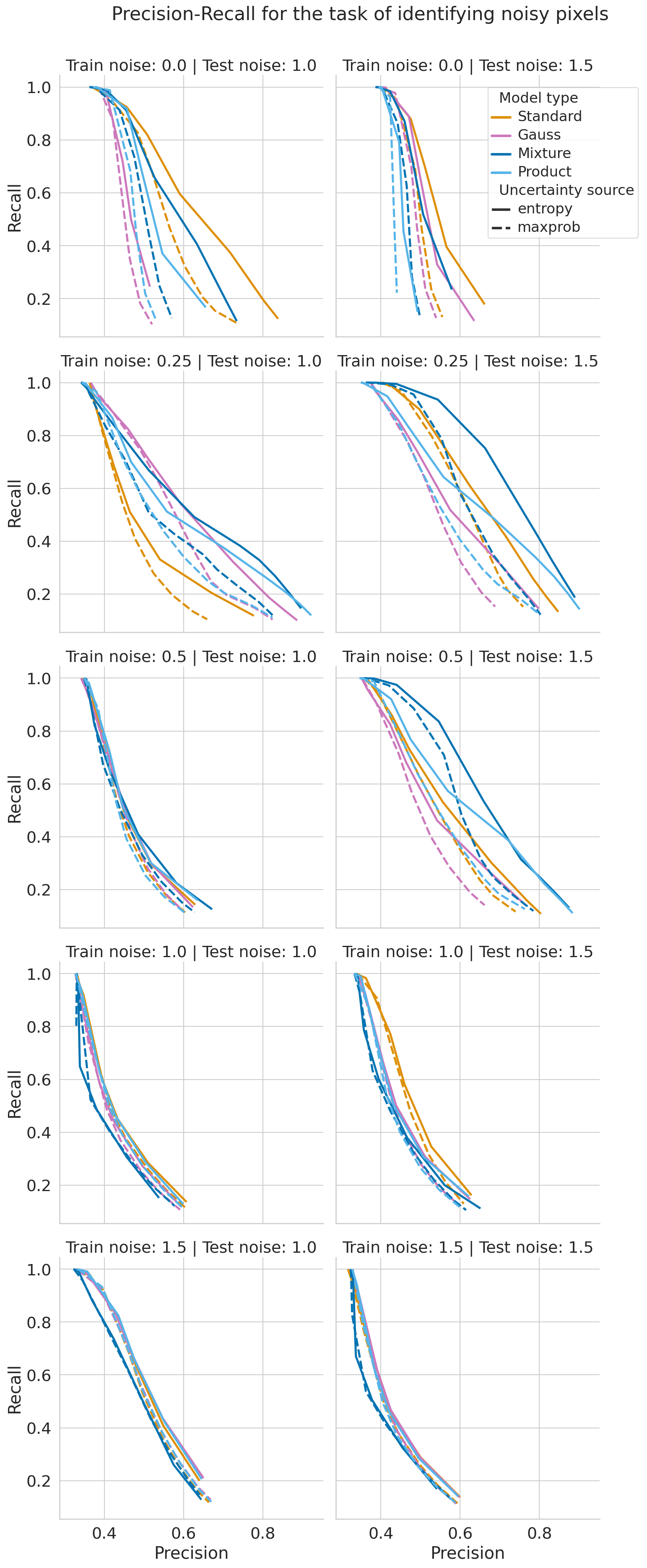}
  \caption{Elliptical noise on PASTIS. See Figure \ref{pastis_elliptic_noise_appendix} in Appendix for further results.}
  \label{pastis_elliptic_noise}
\end{subfigure}
\label{pastis_noise}
\caption{Uncertainty-Noise Precision-Recall curves at different train and test noise levels by model type and uncertainty measure.}
\end{figure}

\begin{figure}[h!]
  \centering
  \includegraphics[width=197.2pt]{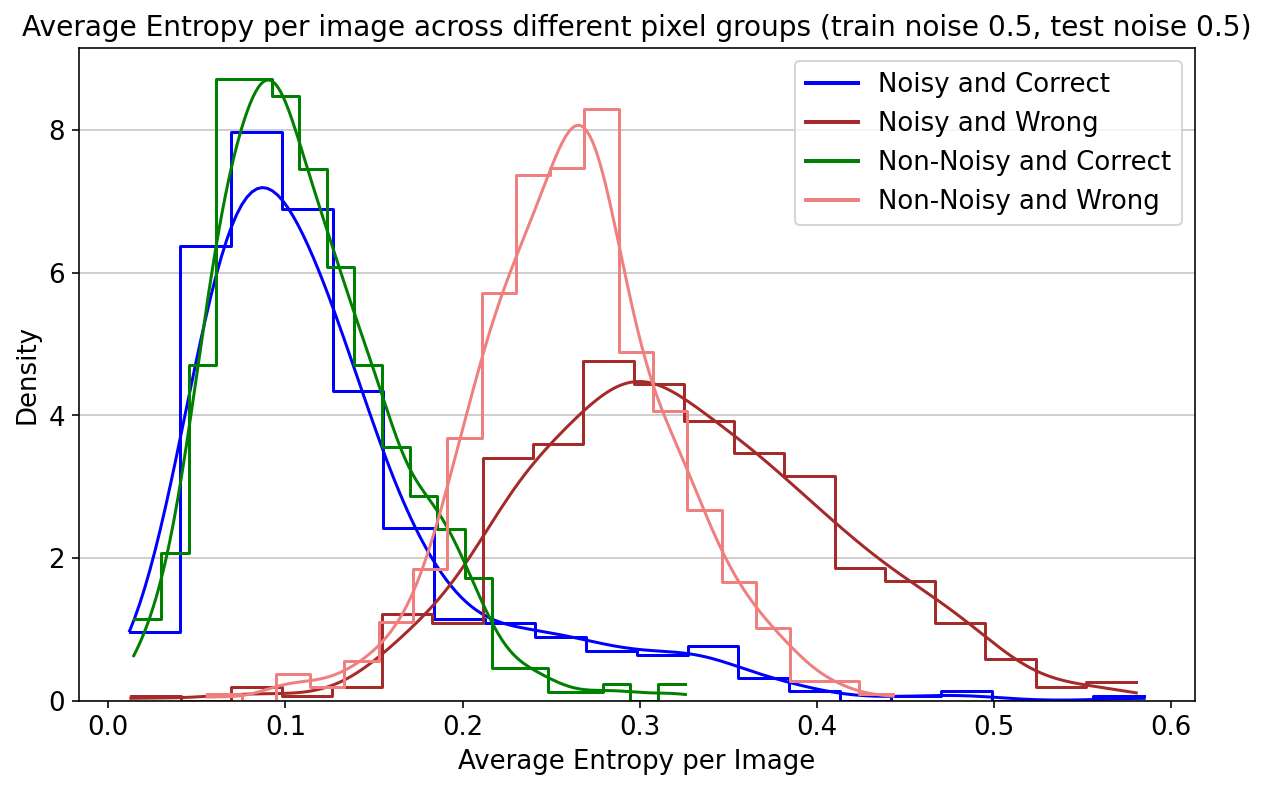}
  \includegraphics[width=197.2pt]{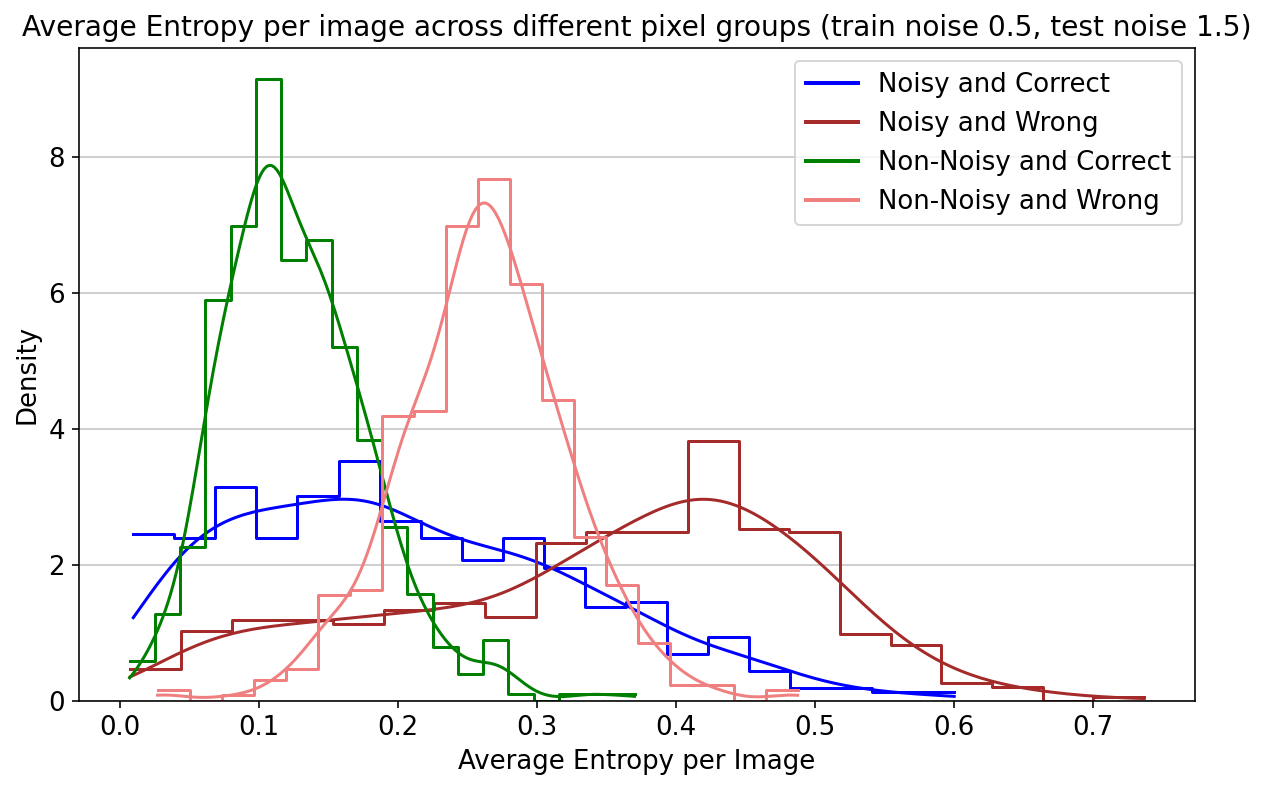}
  \caption{Distribution of average entropy per image computed for different sets of pixels: noise corrupted pixels which were correctly classified, noise corrupted pixels wrongly classified, non-noisy pixels correctly classified, non-noisy pixels wrongly classified. These distributions were computed for the TSViT SSN on the PASTIS test set. (Left): Model trained and evaluated with a noise level 0.5. (Right): Model trained with a noise level 0.5 but evaluated with a higher noise level of 1.5. We observe a clearer difference between distributions in this noise setting.}
  \label{entropy_distribution_under_noise}
\end{figure}

\paragraph{Identifying noisy pixels}
Analysis of the Uncertainty-Noise Precision-Recall curves  for noisy pixel detection in Figures~\ref{pastis_object_noise} and \ref{pastis_elliptic_noise} reveals several patterns in how the  uncertainty metrics, model types, and noise levels affect performance. First, for any given model, entropy consistently outperforms the maxprob score as the uncertainty metric for this task. This is consistent with the findings in \citep{blum2019fishyscapes} who observed that the maxprob score performed poorly at OOD detection. The ability to identify noisy pixels, however, is highly dependent on the relative noise levels during training and testing. Models struggle to detect noisy regions when trained and tested on data with the same level of noise: models adapt to that level of noise and the distribution of uncertainty estimates on noisy pixels approaches that of uncorrupted pixels as shown in Figure~\ref{entropy_distribution_under_noise}.  Similarly, when the training noise level is higher than the test noise level, differentiating between noisy and ``clean'' regions remains challenging.
Effective noise identification is only achieved when the test noise level is substantially higher than the training one. In this regime, ensemble models generally perform best, with ensemble mixtures exhibiting a slight edge.

\begin{figure}[h]
  \centering
  \includegraphics[width=\columnwidth]{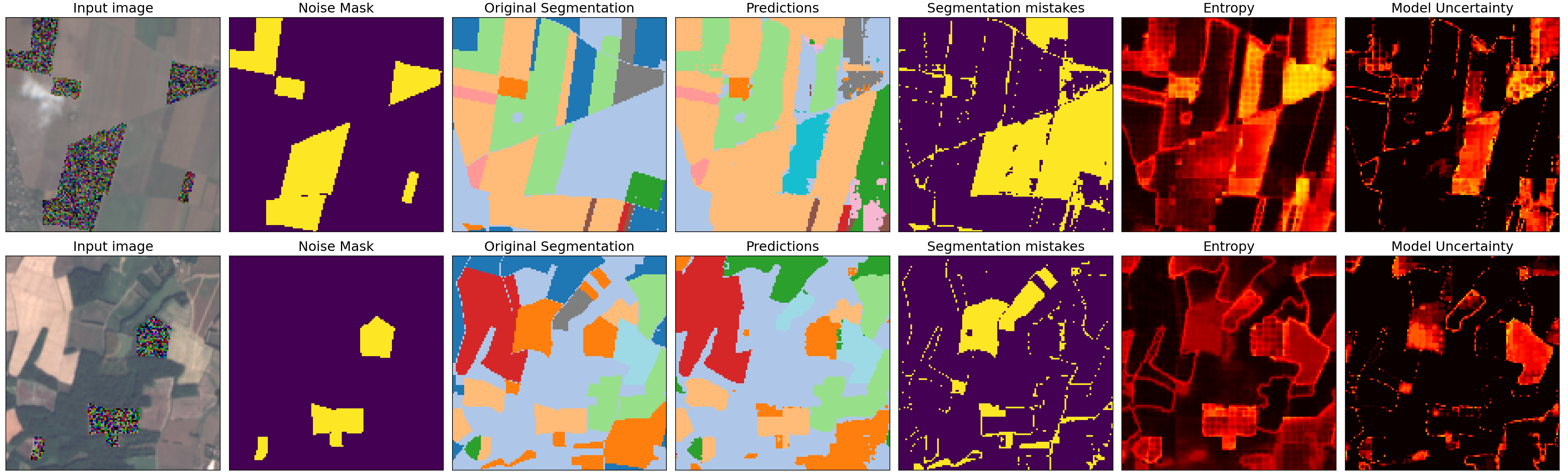}
  \includegraphics[width=\columnwidth]{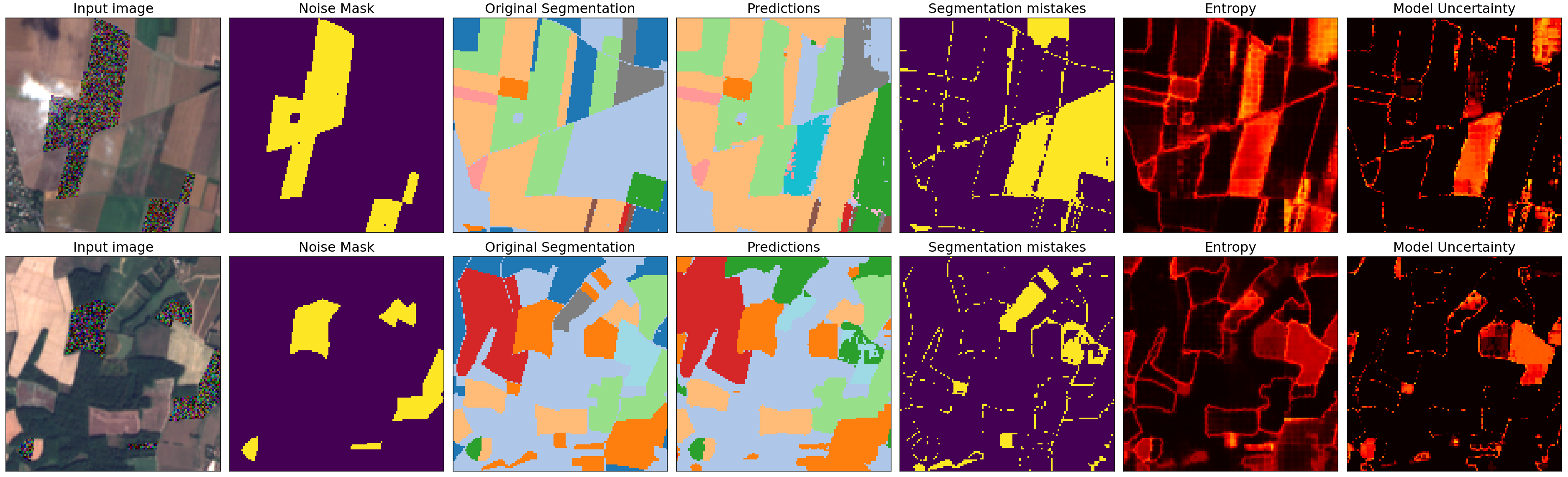}
  \caption{Noise corrupted examples from PASTIS using Object-Level noise and predictions obtained from a SSN model. From left to right: input image with added Gaussian noise, noise mask indicating which pixels have been corrupted, segmentation labels, segmentation predicted by the model, mask indicating segmentation mistakes, entropy heatmap, model uncertainty heatmap (sampled categorical variation) indicating the variety of alternative predictions at each pixel. The top two rows show results for a model trained on uncorrupted data, whereas for the bottom two rows we used a model trained on the same noise level as used as test time (1.5). We can see that for the model trained without noise, some noisy regions lit up in the entropy heatmap even if correctly predicted, and this effect largely disappears for the model trained with noise. Note that the "void" class is depicted using dark blue segments and is ignored during training. }
  \label{pastis_examples}
\end{figure}

\FloatBarrier

\subsection{ForTy}
 
\begin{figure}[h!]
  \centering
  \includegraphics[width=0.7\textwidth]{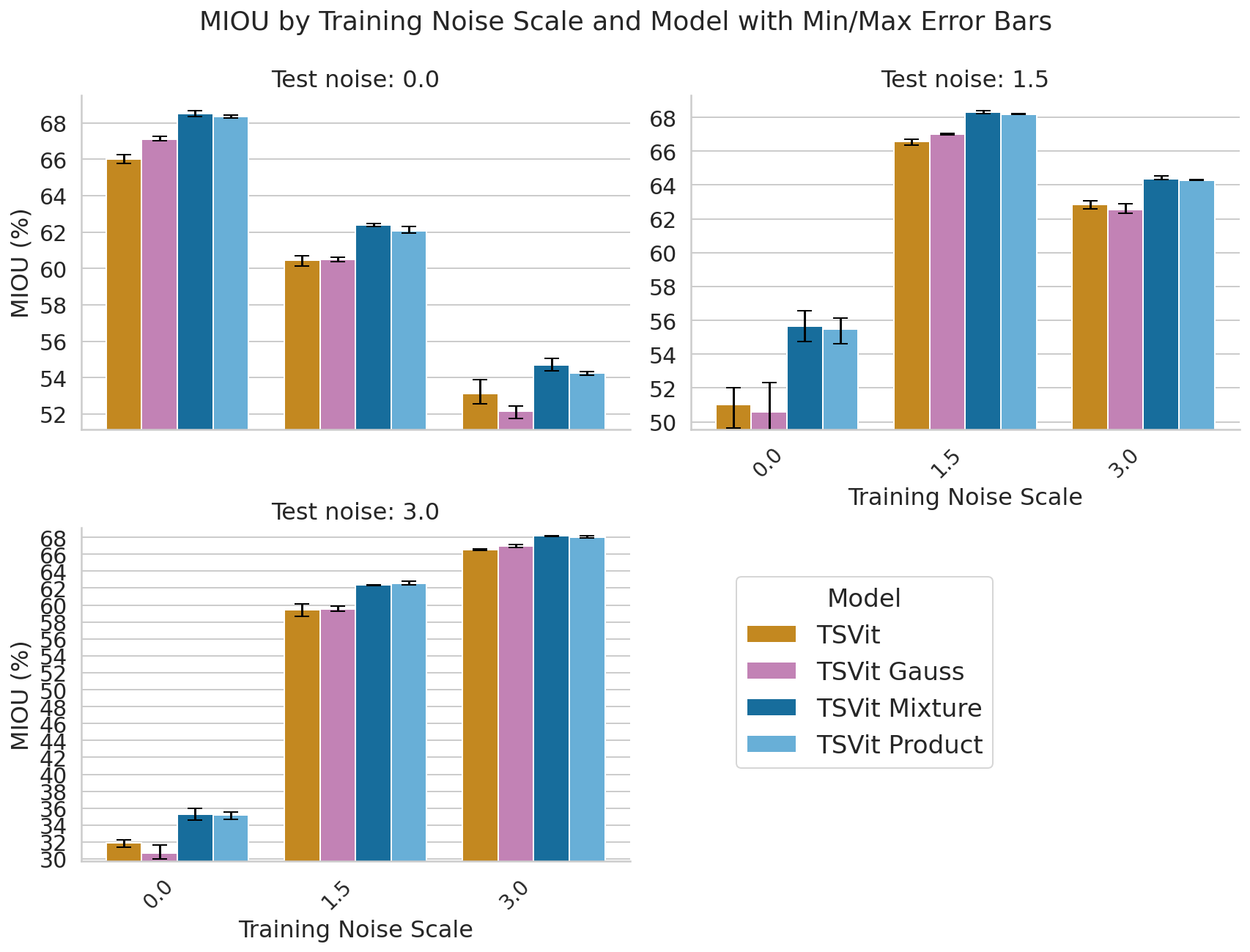}
  \caption{Elliptical noise on ForTy: Segmentation performance by model on different train/test noise level combinations, using random shape noise. }
  \label{forty_noise_miou}
\end{figure}

\paragraph{Segmentation performance} 
The segmentation results on ForTy shown in Figure~\ref{forty_noise_miou}, exhibit a similar pattern to those of the PASTIS experiments: ensembles consistently perform best, and the SSNs outperform the standard models when training with an adequate noise level. Notably, models trained on the ForTy dataset tolerated significantly higher noise levels before a substantial drop in performance occurred, which is why the highest noise level we used for this dataset was 3. We attribute this enhanced robustness to two primary factors: dataset scale and data structure. First, ForTy is a considerably larger dataset than PASTIS. Second, the datasets possess distinct structural characteristics. PASTIS is focused on French agricultural land, which primarily consists of geometrically regular parcels with well-defined borders, while ForTy is a global dataset of forest types, characterized by more amorphous regions and a highly heterogeneous distribution of segment sizes, including both small patches and vast, contiguous areas of a single class. We conjecture that this structural disparity influences the model's response to noise. In large, homogeneous regions typical of ForTy, the model can effectively learn to ``in-paint'' or ignore noisy regions by relying on strong contextual cues. Conversely, in regions dense with small, intricate objects (where the classification task is inherently more complex) the model's performance is more susceptible to degradation from such noise corruption. This effect is illustrated in Figure \ref{forty_examples} and more examples can be found in the Appendix \ref{forty_ex_ellipse_1}, \ref{forty_ex_ellipse_2}.

\begin{figure}[h!]
  \centering
  \includegraphics[width=0.65\textwidth]{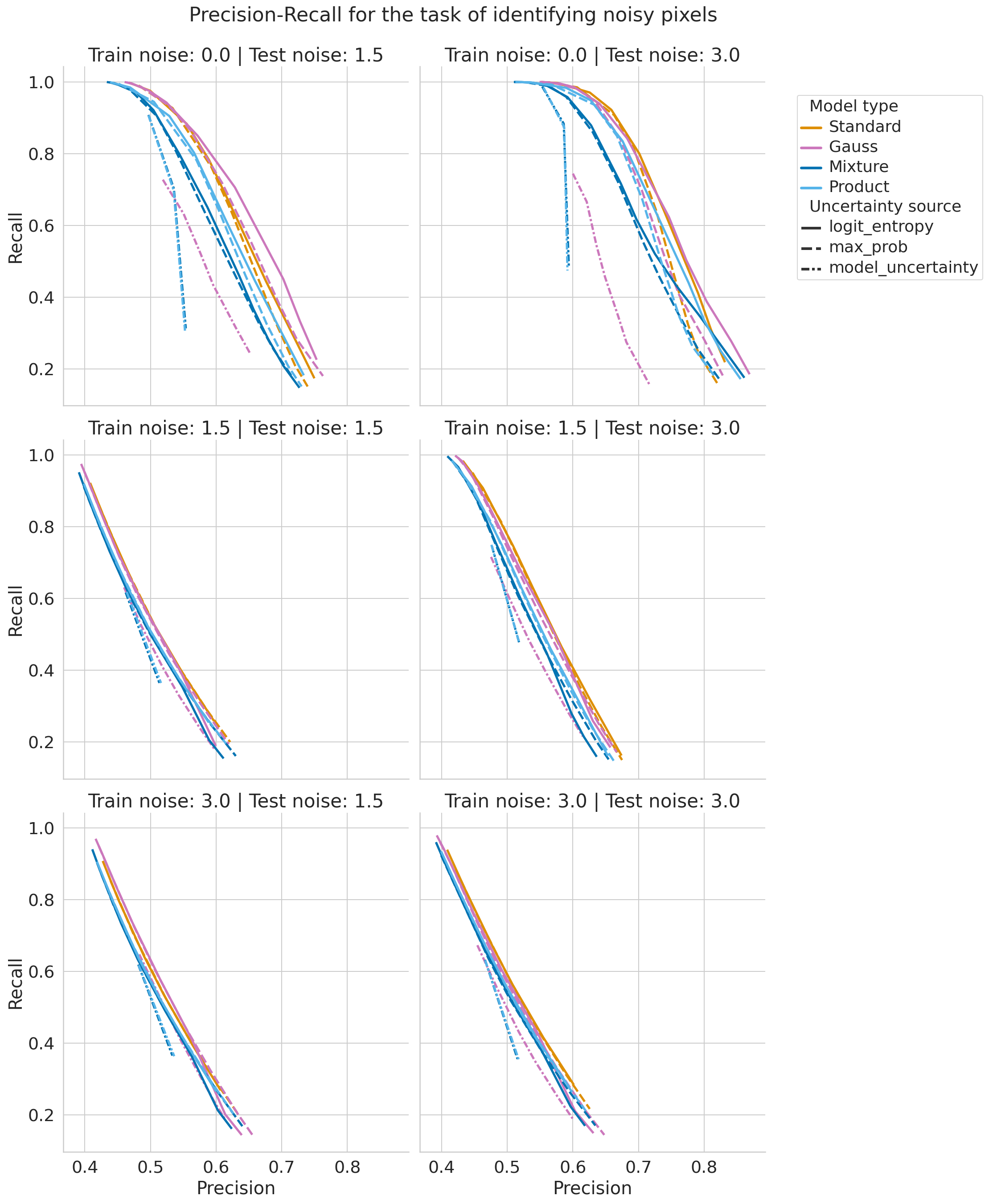}
  \caption{Elliptical noise on ForTy: Uncertainty-Noise Precision-Recall curves at different train and test noise levels by model type and uncertainty source. }
  \label{fig:forty_noisy_pr_curves}
\end{figure}

\paragraph{Identifying noisy pixels} Figure \ref{fig:forty_noisy_pr_curves} shows the Uncertainty-Noise Precision-Recall curves for identifying noisy pixels generated using elliptical noise described in Section~\ref{sec:noise_corruption}. Two main patterns in these results mirror those on PASTIS: Models can identify noisy pixels effectively only when the training noise level is much lower than the test noise level; and entropy is the best performing uncertainty measure on this task. Unlike on PASTIS, however, where ensemble models performed best, here SSN is the best-performing model class, followed by the standard models, delegating ensembles to the third place.

\begin{figure}[h!]
  \centering
  \includegraphics[width=1.0\textwidth]{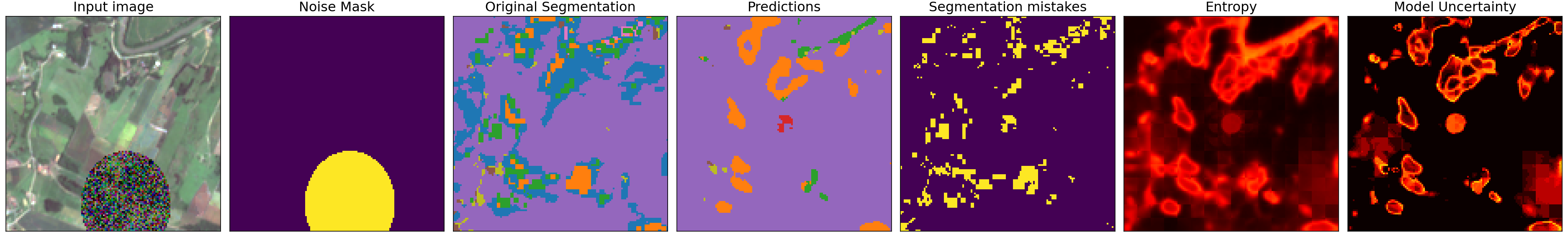}
  \includegraphics[width=1.0\textwidth]{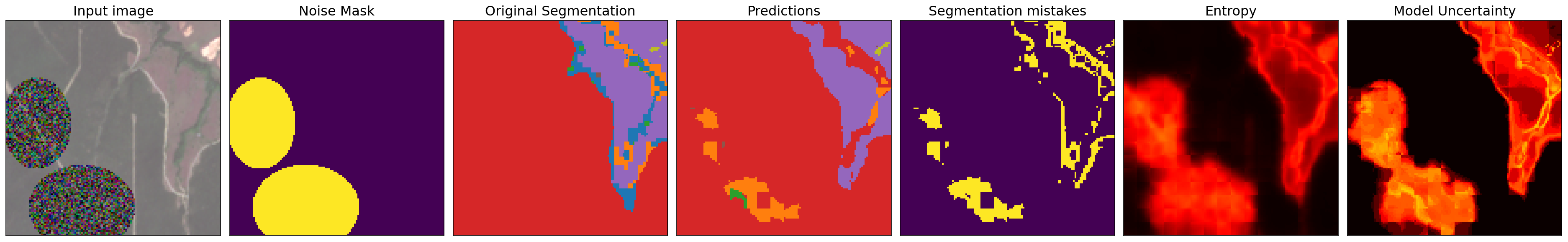}
  \caption{Examples from ForTy corrupted with elliptical noise, and  the corresponding predictions from a SSN model. From left to right: input image with added Gaussian noise, noise mask indicating which pixels have been corrupted, segmentation labels, segmentation predicted by the model, mask indicating segmentation mistakes, entropy heatmap, model uncertainty heatmap indicating the variety of alternative predictions at each pixel (Sampled Categorical Variation). In the bottom row, noise is added to a fairly uniform region of the input and the corresponding pixels lit up clearly in the entropy heatmap. In the top row, noise is added to a region with a more complex underlying structure as can be seen in the segmentation mask and the noisy region cannot be identified from the entropy heatmap. Further examples can be found in the Appendix \ref{forty_ex_ellipse_1}, \ref{forty_ex_ellipse_2}.}
  \label{forty_examples}
\end{figure}

\FloatBarrier

\section{Conclusion}

Quantifying the uncertainty of deep learning models is a complex challenge, as there is no directly accessible ``ground truth'' for the uncertainty itself. This makes rigorous evaluation paramount. Relying solely on qualitative assessments, such as visualizing uncertainty maps for semantic segmentation, can be misleading and risks overestimating the reliability of these estimates. We argue that the most effective evaluation strategy is to assess uncertainty through its utility for carefully chosen and well-defined downstream tasks. By defining what we expect an uncertainty estimate to be useful for, for example identifying misclassified pixels or flagging particularly challenging input images, we can create concrete benchmarks to evaluate and compare different uncertainty estimation methods robustly.

Our experiments lead to several practical takeaways. The strong performance and inherent robustness of Transformer architectures make them a solid foundation for tasks requiring dependable uncertainty estimates. When computational resources permit, ensembles continue to provide best segmentation accuracy and uncertainty estimation. However, Stochastic Segmentation Networks (SSNs) present a compelling alternative, striking an excellent balance between predictive performance, model efficiency, and resilience to input noise. As SSNs model the dependence between the labels at different spatial locations, they might capture additional uncertainty information not available to models that do not capture this dependence. Testing whether this is indeed the case and how to take advantage of such richer uncertainty information is an interesting direction for future work.

Given the significant performance variations we observed across different tasks and datasets, we strongly advocate for a standardized evaluation protocol. We recommend that practitioners evaluate uncertainty estimates on a dedicated held-out test set, to allow for a more targeted and rigorous assessment of a model's predictive uncertainty before deployment. Our findings also confirm that image-level uncertainty metrics correlate well with overall segmentation quality.

This work opens several avenues for future research. The difficulty in pinpointing pixel-level errors suggests that exploring uncertainty at intermediate granularities, such as superpixels, could be a promising direction. Key questions would then be how to define the optimal scale for these regions and how to aggregate uncertainty within them effectively. Furthermore, investigating other ensemble methods such as MC Dropout or exploring ways to explicitly increase the diversity within ensembles could bring potential improvements. Finally, taking into account the impact of different loss functions (e.g., Dice or Focal loss) on uncertainty calibration could lead to improved architecture-loss combinations.

\begin{ack}
We thank Vaibhav Rajan for helpful comments and suggestions.
\end{ack}

{
    \small
    \bibliographystyle{plainnat}
    \bibliography{main}
}

%%%%%%%%%%%%%%%%%%%%%%%%%%%%%%%%%%%%%%%%%%%%%%%%%%%%%%%%%%%%
\newpage

\appendix

\section{Appendix}

\subsection{Hyperparameters} \label{hyper_appendix}

For all experiments using Stochastic Segmentation Networks, we use 32 samples during training and 16 samples for predictions. The rank of the covariance matrix was fixed to 10 throughout. The scaling factors for the matrices  $P, D$ were set to 0.01 for UNET3D, 0.05 for UTAE and 0.05 for TSViT. 

\begin{table}[h!]
\centering
\caption{TSViT Hyperparameter Comparison for PASTIS and ForTy Datasets}
\label{tab:hyperparams}
\begin{tabular}{@{}lcc@{}}
\toprule
\textbf{Hyperparameter} & \textbf{PASTIS} & \textbf{Forty} \\
\midrule
\texttt{patch\_sizes} & (1, 4, 4) & (1, 8, 8)\\
\texttt{decoder\_depth} & 2 & 2 \\
\texttt{temporal\_depth} & 4 & 2 \\
\texttt{spatial\_depth} & 4 & 2 \\
\texttt{num\_heads} & 4 & 6 \\
\texttt{emb\_dim} & 192 & 192 \\
\bottomrule
\end{tabular}
\end{table}

\FloatBarrier

\subsection{Results on uncorrupted data}

\begin{figure}[h!]
  \centering
  \includegraphics[width=270pt]{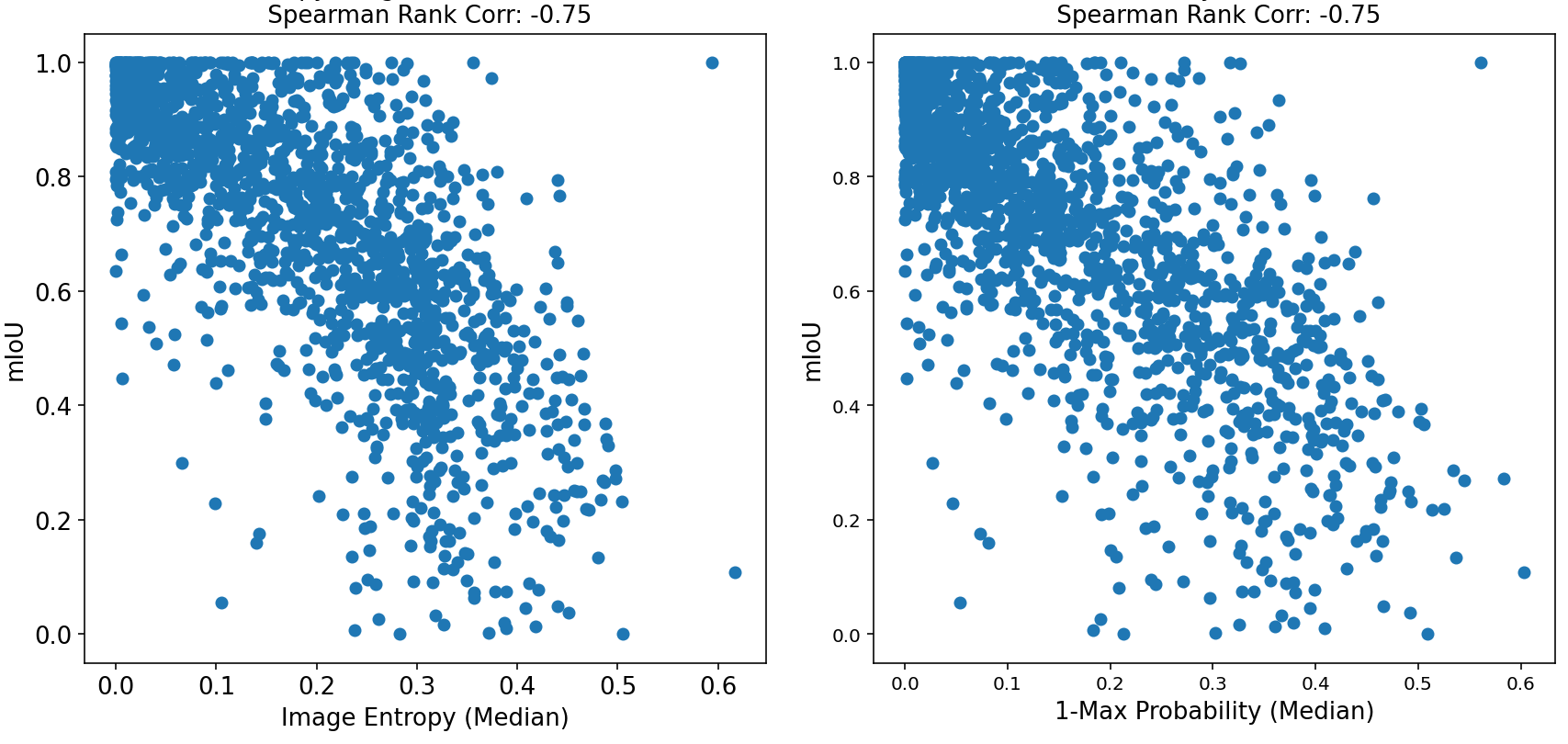}
  \caption{Image mIoU vs uncertainty measures for TSViT SSN on 1600 randomly drawn images from ForTy test set, Two different uncertainty estimation are used. Left: Median pixel entropy over the image. Right: Median 1-max probability over the image.}
  \label{forty_scatter_median_appendix}
\end{figure}

\begin{figure}[h!]
  \centering
  \includegraphics[width=\columnwidth]{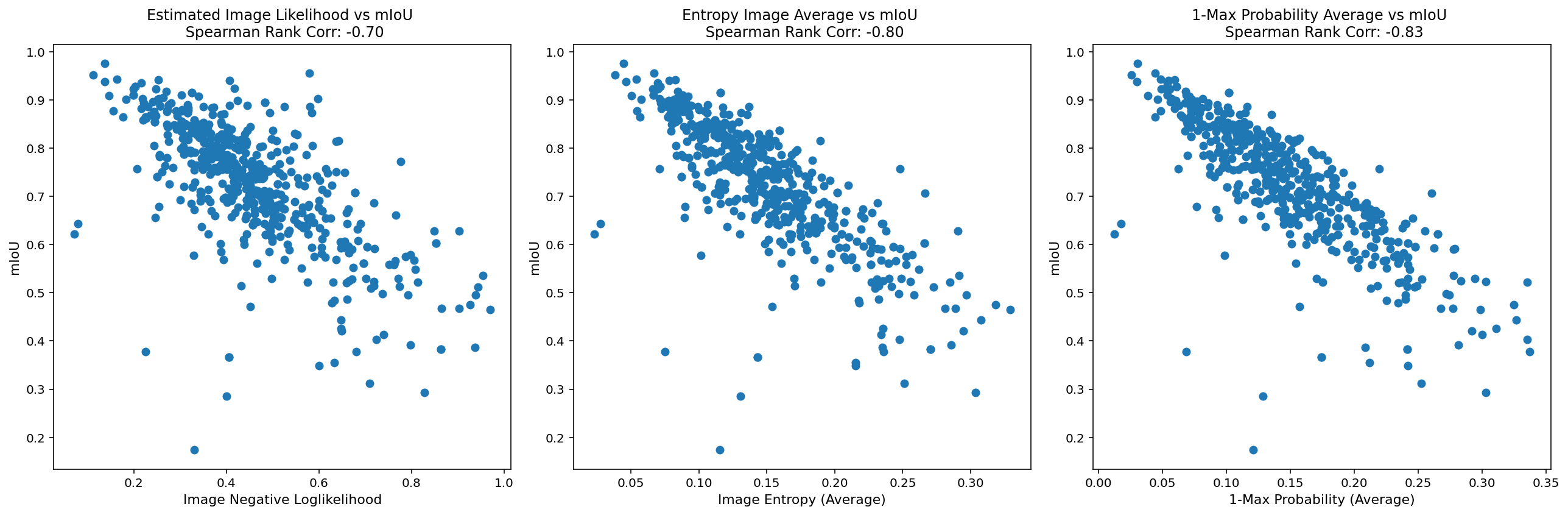}
  \caption{Image mIoU vs.~uncertainty measures for UNET3D SSN on the PASTIS test set. Three measures of  uncertainty were considered: Left: Negative log-likelihood for the entire image estimated using 32 latent samples. Middle: (Marginal) pixel entropy averaged over the image. Right: 1-max probability averaged over the image.}
  \label{pastis_scatter_unet3d_appendix}
\end{figure}

\FloatBarrier
\newpage

\subsection{Results on noise corrupted data}

\begin{figure}[h!]
  \centering
  \includegraphics[width=1.1\textwidth]{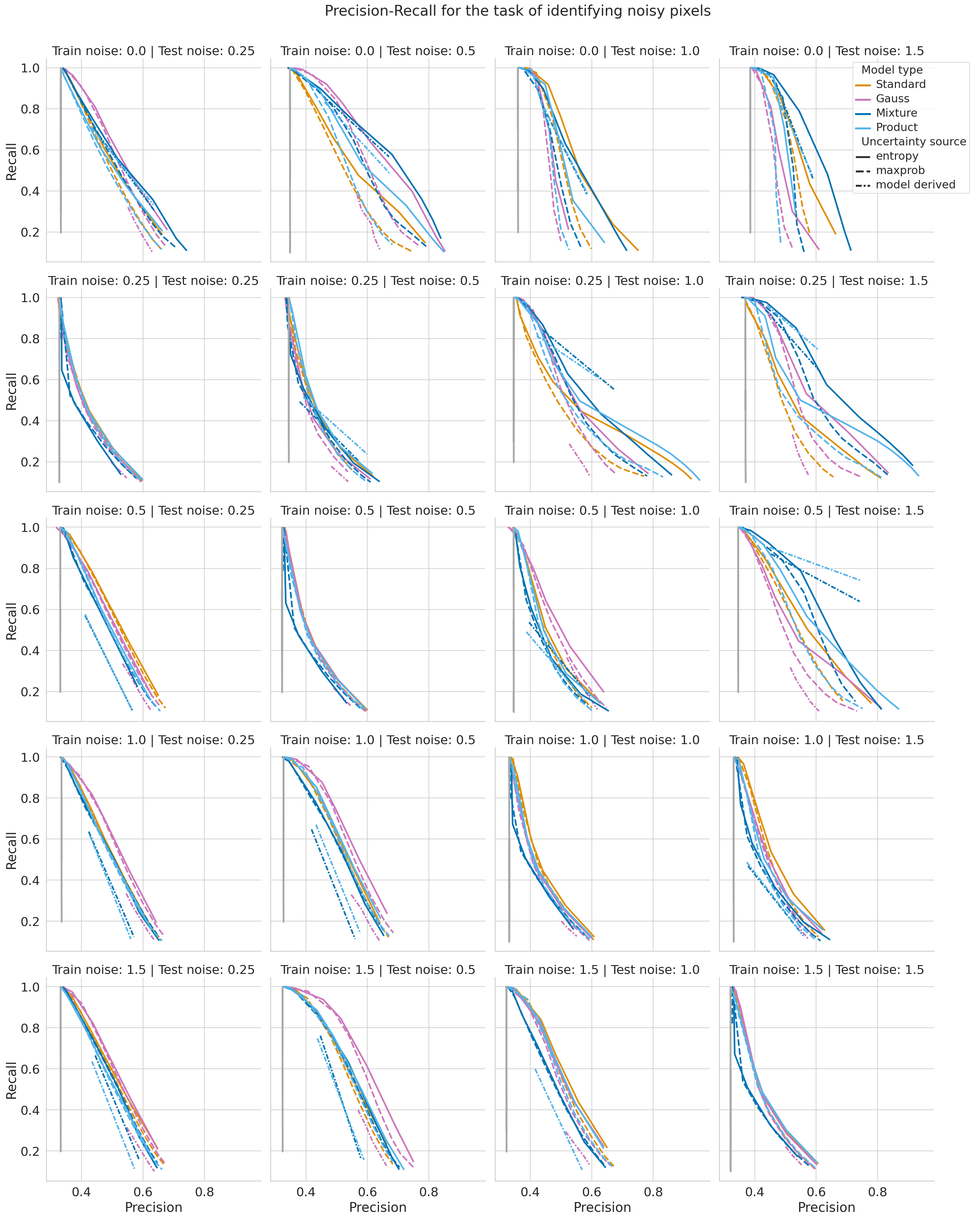}
  \caption{Object level noise on PASTIS: Precision-Recall curves at different train and test noise levels by model type and uncertainty source. The vertical gray line is the baseline obtained by sampling for each pixel independently a random uniform uncertainty value in $[0, 1]$. Note that the resulting precision value is higher than the average proportion of noisy pixels $(\sim0.18)$ since we exclude uncorrupted, incorrectly classified (in the original segmentation task) pixels for which the model predicts high uncertainty as explained in Section \ref{sec:assessing_uncertainty}.}
  \label{pastis_object_noise_appendix}
\end{figure}

\begin{figure}[h!]
  \centering
  \includegraphics[width=1.1\textwidth]{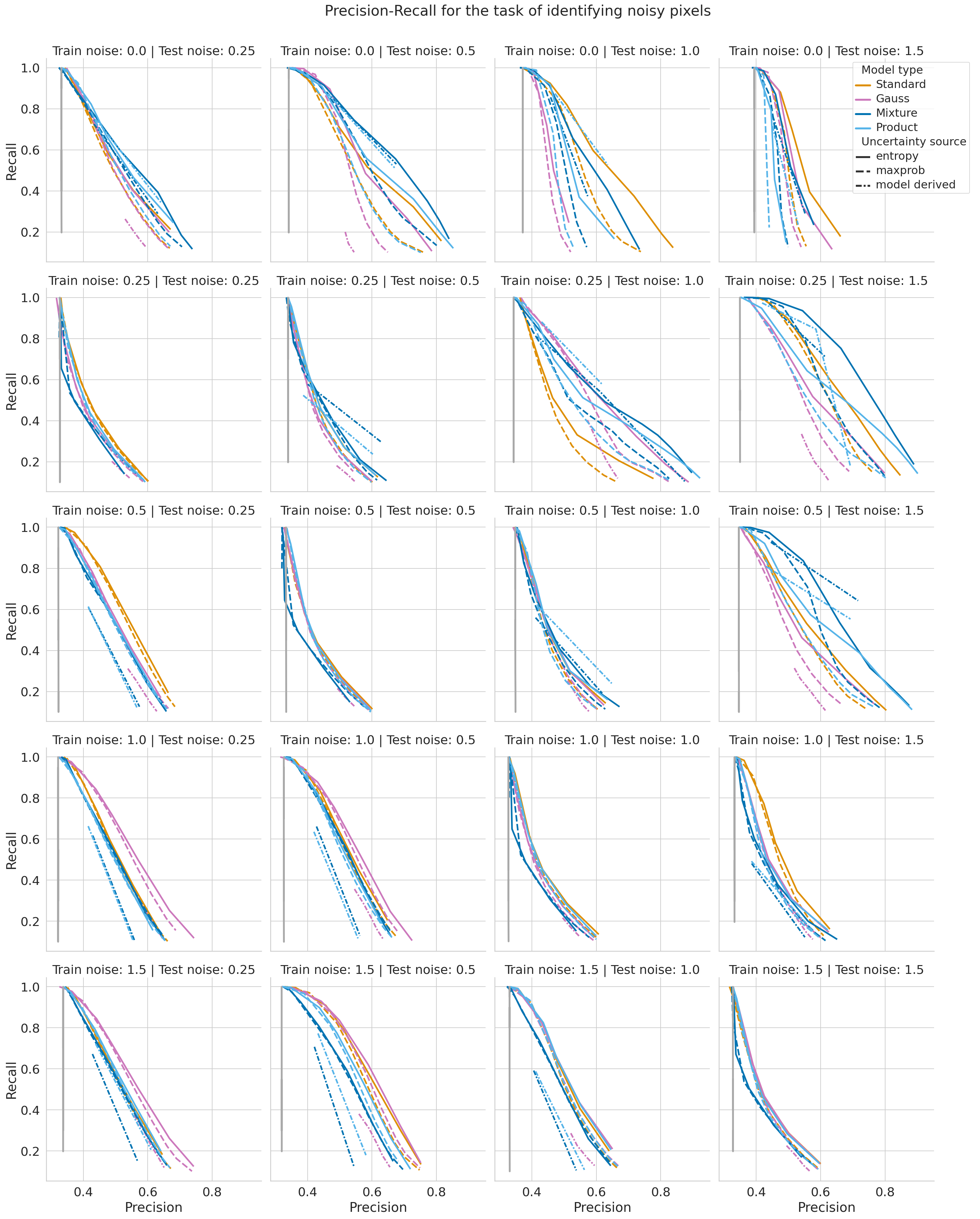}
  \caption{Elliptical noise on PASTIS: Precision-Recall curves at different train and test noise levels by model type and uncertainty source.The vertical gray line is the baseline obtained by sampling for each pixel independently a random uniform uncertainty value in $[0, 1]$. Note that the resulting precision value is higher than the average proportion of noisy pixels $(\sim0.16)$ since we exclude uncorrupted, incorrectly classified (in the original segmentation task) pixels for which the model predicts high uncertainty as explained in Section \ref{sec:assessing_uncertainty}.}
  \label{pastis_elliptic_noise_appendix}
\end{figure}

\begin{figure}[h!]
  \centering
  \includegraphics[width=\columnwidth]{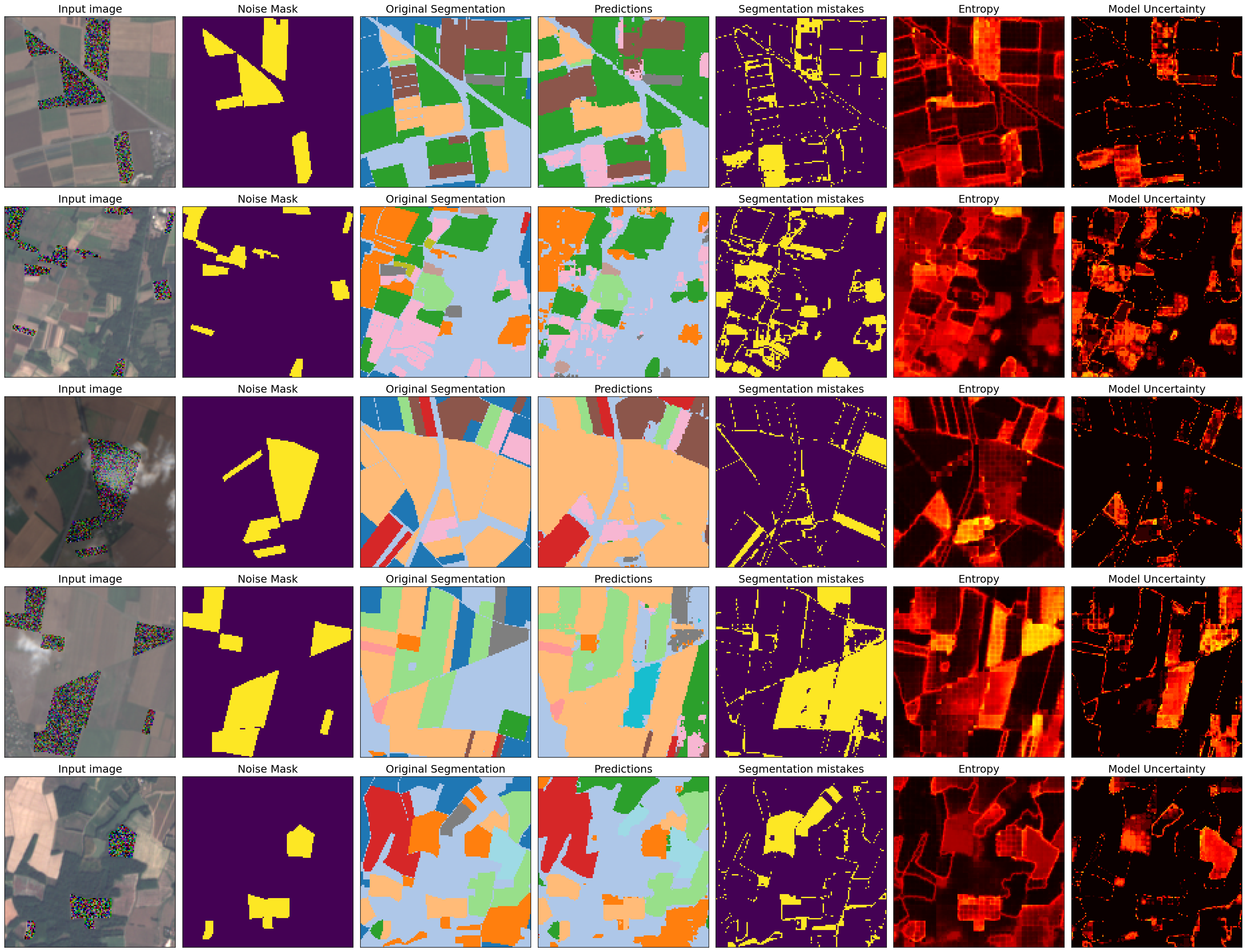}
 \caption{Noise corrupted examples from PASTIS using Object-Level noise and predictions obtained from a SSN model. From left to right: input image with added Gaussian noise, noise mask indicating which pixels have been corrupted, segmentation labels, segmentation predicted by the model, mask indicating segmentation mistakes, entropy heatmap, model uncertainty heatmap indicating the variety of alternative predictions at each pixel. Results are shown for a model trained on uncorrupted data. We can see that some noisy regions lit up in the entropy heatmap even if correctly predicted. Note that the "void" class is depicted using dark blue segments and is ignored during training. }
  \label{gauss_tsvit_noise_examples}
\end{figure}

\begin{figure}[h!]
  \centering
  \includegraphics[width=\columnwidth]{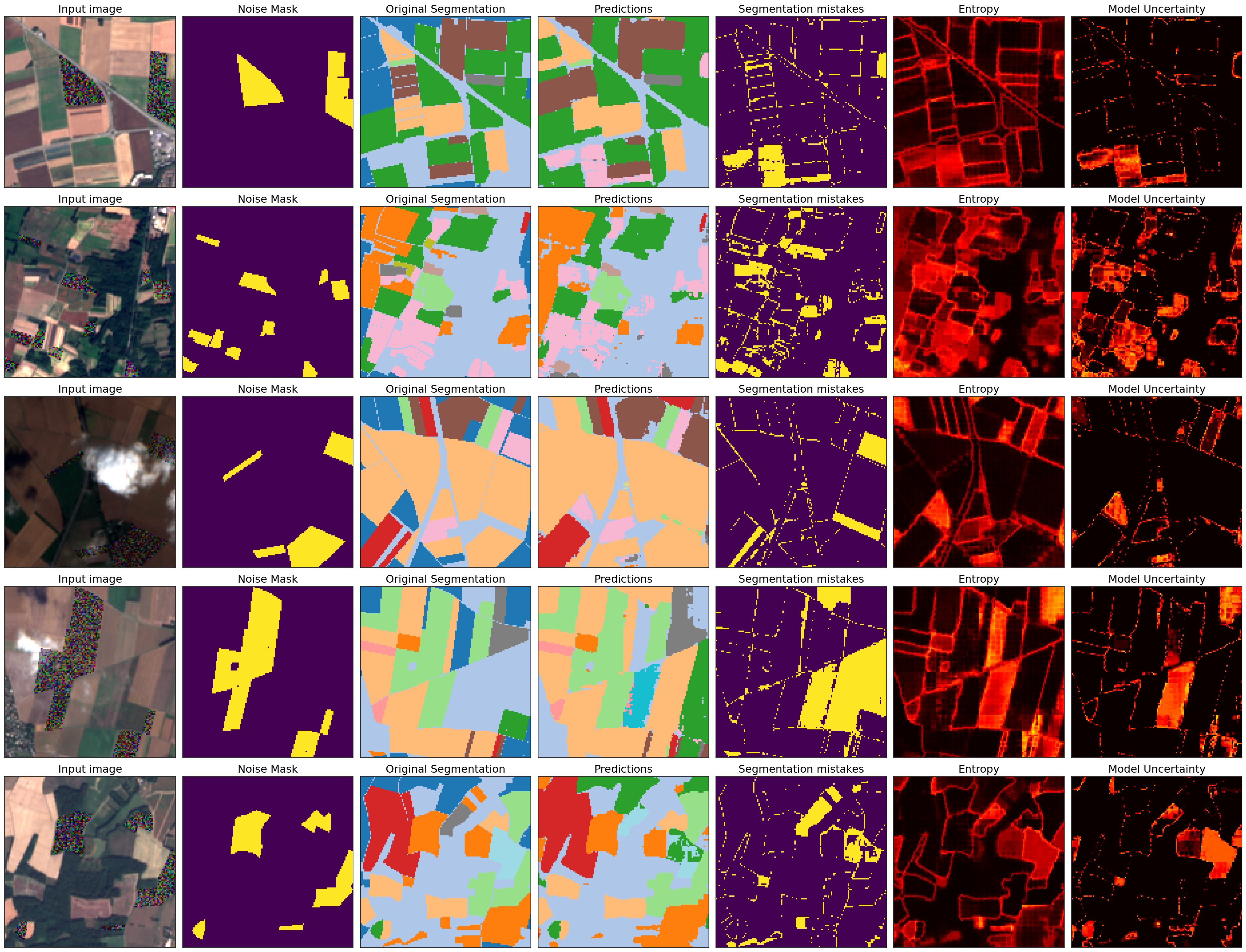}
  \caption{Noise corrupted examples from PASTIS using Object-Level noise and predictions obtained from a SSN model. From left to right: input image with added Gaussian noise, noise mask indicating which pixels have been corrupted, segmentation labels, segmentation predicted by the model, mask indicating segmentation mistakes, entropy heatmap, model uncertainty heatmap indicating the variety of alternative predictions at each pixel. Results are shown for a model trained with same noise level as used at test time (1.5). We can see that noisy regions in general do not lit up in the entropy heatmap when correctly predicted. Note that the "void" class is depicted using dark blue segments and is ignored during training. }
  \label{gauss_tsvit_noise_examples_2}
\end{figure}

\begin{figure}[h!]
  \centering
  \includegraphics[width=\columnwidth]{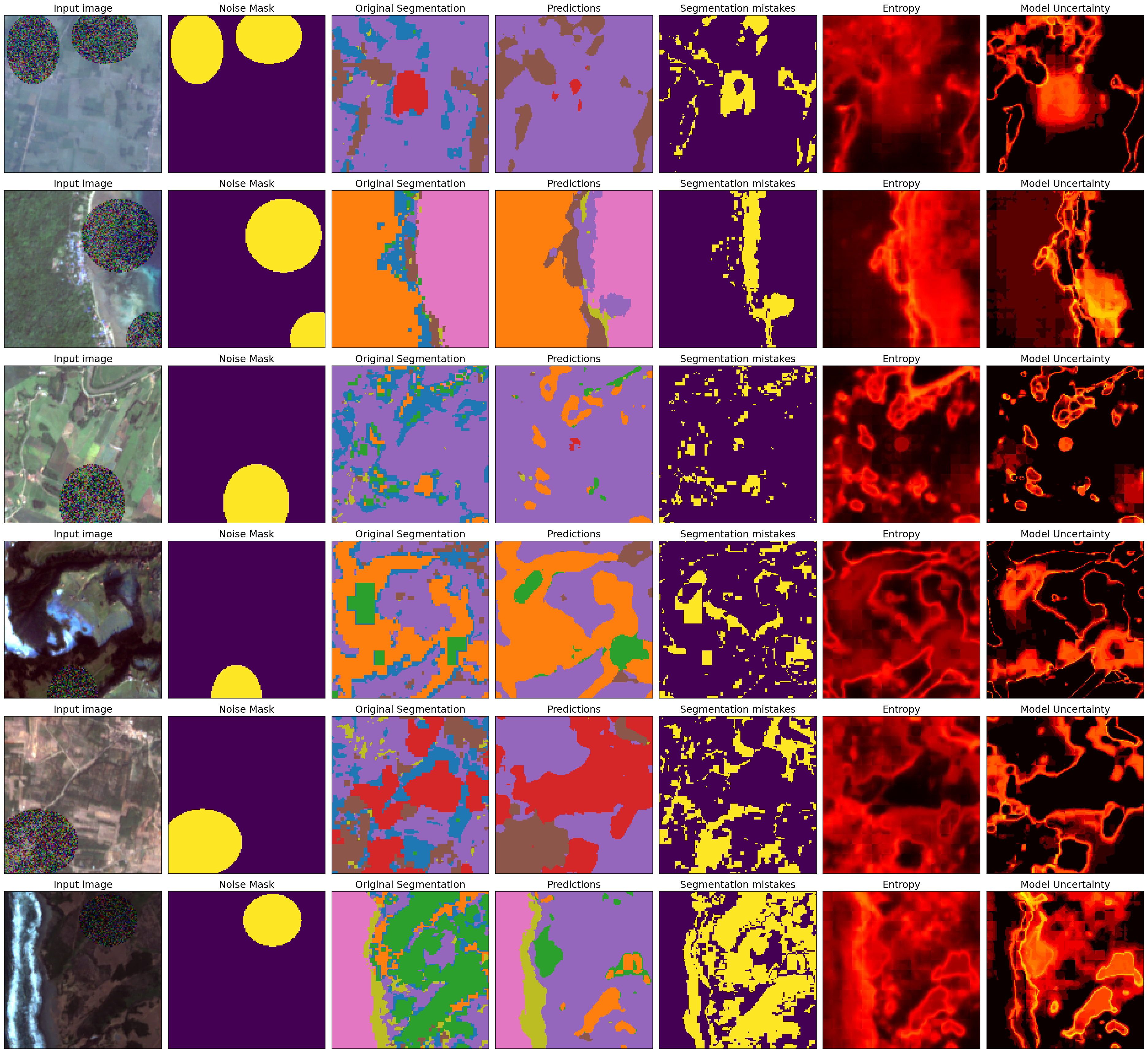}
  \caption{Examples from ForTy corrupted with elliptical noise, and the corresponding predictions from a SSN model. From left to right: input image with added Gaussian noise, noise mask indicating which pixels have been corrupted, segmentation labels, segmentation predicted by the model, mask indicating segmentation mistakes, entropy heatmap, model uncertainty heatmap indicating the variety of alternative predictions at each pixel. In these examples the noise was added to a region with a relatively complex underlying structure as can be seen in the segmentation mask and the noisy region cannot be identified from the entropy heatmap.}
  \label{forty_ex_ellipse_1}
\end{figure}

\begin{figure}[h!]
  \centering
  \includegraphics[width=\columnwidth]{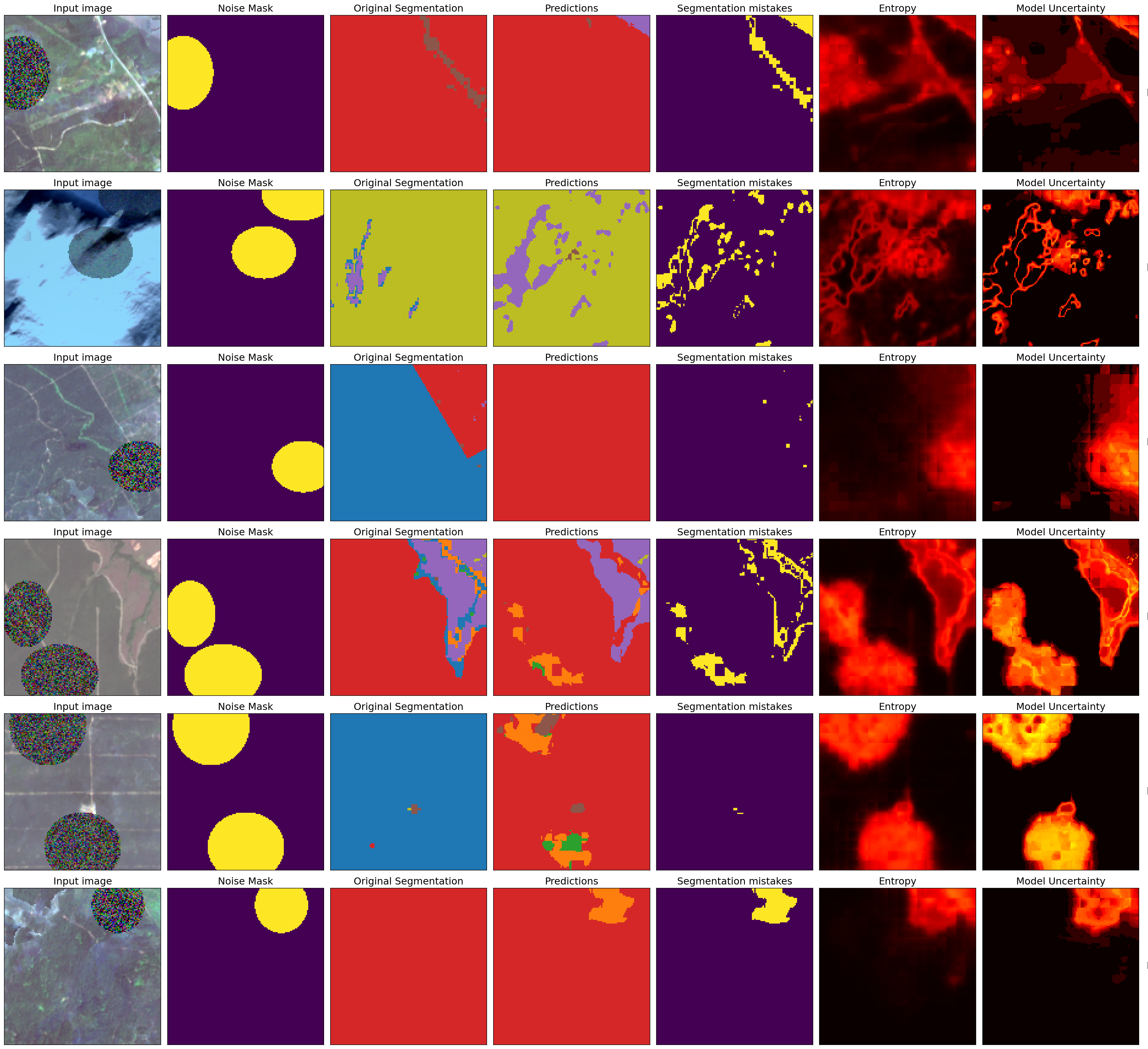}
  \caption{Examples from ForTy corrupted with elliptical noise, and  the corresponding predictions from a SSN model. From left to right: input image with added Gaussian noise, noise mask indicating which pixels have been corrupted, segmentation labels, segmentation predicted by the model, mask indicating segmentation mistakes, entropy heatmap, model uncertainty heatmap indicating the variety of alternative predictions at each pixel. In theses examples the noise was added to a fairly uniform region of the input and the corresponding pixels lit up clearly in the entropy heatmap.}
  \label{forty_ex_ellipse_2}
\end{figure}

\end{document}